%% file: main.tex
\documentclass{article}

\usepackage[final,nonatbib]{neurips_2021}

\usepackage[numbers]{natbib}
\usepackage[utf8]{inputenc}
\usepackage[T1]{fontenc}
\usepackage{url}
\usepackage{booktabs}
\usepackage{amsfonts}
\usepackage{nicefrac}
\usepackage{microtype}
\usepackage[dvipsnames]{xcolor}
\usepackage{graphics}
\usepackage{graphicx}
\usepackage{caption}
\usepackage{subcaption}
\usepackage{colortbl}
\usepackage{wrapfig}
\usepackage{float}
\usepackage{appendix}
\usepackage[format=plain,
            labelfont=it,
            labelsep=period]{caption}
\usepackage{algorithm}
\usepackage{algpseudocode}
\usepackage{amsmath}
\usepackage{amssymb}
\usepackage{mathtools}
\usepackage{amsthm}
\usepackage[symbol]{footmisc}

\usepackage{bbding}
\usepackage{fancyvrb}
\usepackage[scaled=0.85]{DejaVuSansMono}
\usepackage{listings}
\definecolor{codegreen}{rgb}{0,0.5,0}
\definecolor{codered}{rgb}{0.7,0.1,0.1}
\definecolor{codegray}{rgb}{0.5,0.5,0.5}
\definecolor{codepurple}{rgb}{0.58,0,0.82}
\definecolor{backcolour}{rgb}{1,1,1}
\lstdefinestyle{python}{
    language=Python,
    backgroundcolor=\color{backcolour},   
    commentstyle=\color{codered}\textit,
    keywordstyle=\bfseries\color{codegreen},
    numberstyle=\tiny\color{codegray},
    stringstyle=\color{codepurple},
    basicstyle=\ttfamily\footnotesize,
    breakatwhitespace=false,         
    breaklines=true,                 
    captionpos=b,                    
    keepspaces=true,                 
    numbers=left,                    
    numbersep=5pt,                  
    showspaces=false,                
    showstringspaces=false,
    showtabs=false,                  
    tabsize=2,
    fancyvrb=true
}
\newenvironment{codesnippet}
  { \VerbatimEnvironment%
    \begin{Verbatim} }
  { \end{Verbatim}  }  

\theoremstyle{definition}
\newtheorem{definition}{Definition}

\definecolor{red}{rgb/cmyk}{0.9098,0.2471,0.2824 / 0,0.86,0.65,0}
\definecolor{citecolor}{HTML}{0071bc}

\usepackage{hyperref}
\hypersetup{pagebackref=true,breaklinks=true,colorlinks,linkcolor=BrickRed,citecolor=citecolor,urlcolor=citecolor,bookmarks=false}

\title{Stabilizing Deep $Q$-Learning with ConvNets and Vision Transformers under Data Augmentation}

\author{%
  Nicklas Hansen$^{1}$~~~~
  Hao Su$^{1}$~~~~
  Xiaolong Wang$^{1}$\\
  $^{1}$University of California, San Diego\\
  \texttt{nihansen@ucsd.edu~\{haosu,xiw012\}@eng.ucsd.edu}
}

\begin{document}

\maketitle

\begin{abstract}
While agents trained by Reinforcement Learning (RL) can solve increasingly challenging tasks directly from visual observations, generalizing learned skills to novel environments remains very challenging. Extensive use of data augmentation is a promising technique for improving generalization in RL, but it is often found to decrease sample efficiency and can even lead to divergence. In this paper, we investigate causes of instability when using data augmentation in common off-policy RL algorithms. We identify two problems, both rooted in high-variance $Q$-targets. Based on our findings, we propose a simple yet effective technique for stabilizing this class of algorithms under augmentation. We perform extensive empirical evaluation of image-based RL using both ConvNets and Vision Transformers (ViT) on a family of benchmarks based on DeepMind Control Suite, as well as in robotic manipulation tasks. Our method greatly improves stability and sample efficiency of ConvNets under augmentation, and achieves generalization results competitive with state-of-the-art methods for image-based RL in environments with unseen visuals. We further show that our method scales to RL with ViT-based architectures, and that data augmentation may be especially important in this setting.\footnote[2]{Website and code is available at: \url{https://nicklashansen.github.io/SVEA}.}
\end{abstract}

\section{Introduction}
\label{sec:introduction}
\vspace{-0.05in}
Reinforcement Learning (RL) from visual observations has achieved tremendous success in various applications such as video-games \citep{mnih2013playing, Berner2019Dota2W, Vinyals2019GrandmasterLI}, robotic manipulation \citep{levine2016end}, and autonomous navigation \citep{Mirowski2017LearningTN, zhu2017target}. However, it is still very challenging for current methods to generalize the learned skills to novel environments, and policies trained by RL can easily overfit to the training environment \citep{Zhang2018ASO, Farebrother2018GeneralizationAR}, especially for high-dimensional observation spaces such as images~\citep{Cobbe2019QuantifyingGI, Song2020ObservationalOI}.

Increasing the variability in training data via domain randomization~\citep{Tobin_2017,pinto2017asymmetric} and data augmentation~\citep{Shorten2019ASO,laskin2020reinforcement,kostrikov2020image,Raileanu2020AutomaticDA} has demonstrated encouraging results for learning policies invariant to changes in environment observations. Specifically, recent works on data augmentation~\citep{laskin2020reinforcement,kostrikov2020image} both show improvements in sample efficiency from simple cropping and translation augmentations, but the studies also conclude that additional data augmentation in fact \textit{decrease} sample efficiency and even cause divergence. While these augmentations have the potential to improve generalization, the increasingly varied data makes the optimization more challenging and risks instability. Unlike supervised learning, balancing the trade-off between stability and generalization in RL requires substantial trial and error.

In this paper, we illuminate causes of instability when applying data augmentation to common off-policy RL algorithms \citep{mnih2013playing, Lillicrap2016ContinuousCW, Fujimoto2018AddressingFA, haarnoja2018soft}. Based on our findings, we provide an intuitive method for stabilizing this class of algorithms under use of strong data augmentation. Specifically, we find two main causes of instability in previous work's application of data augmentation: (i) indiscriminate application of data augmentation resulting in high-variance $Q$-targets; and (ii) that $Q$-value estimation strictly from augmented data results in over-regularization.

To address these problems, we propose \textbf{SVEA}: \textbf{S}tabilized $Q$-\textbf{V}alue \textbf{E}stimation under \textbf{A}ugmentation, a simple yet effective framework for data augmentation in off-policy RL that greatly improves stability of $Q$-value estimation. Our method consists of the following three components: Firstly, by only applying augmentation in $Q$-value estimation of the \textit{current} state, \textit{without} augmenting $Q$-targets used for bootstrapping, SVEA circumvents erroneous bootstrapping caused by data augmentation; Secondly, we formulate a modified $Q$-objective that optimizes $Q$-value estimation jointly over both augmented and unaugmented copies of the observations; Lastly, for SVEA implemented with an actor-critic algorithm, we optimize the actor strictly on unaugmented data, and instead learn a generalizable policy indirectly through parameter-sharing. Our framework can be implemented efficiently without additional forward passes nor introducing additional learnable parameters.

We perform extensive empirical evaluation on the DeepMind Control Suite \citep{deepmindcontrolsuite2018} and extensions of it, including the DMControl Generalization Benchmark \citep{hansen2021softda} and the Distracting Control Suite \citep{Stone2021TheDC}, as well as a set of robotic manipulation tasks. Our method greatly improve $Q$-value estimation with ConvNets under a set of strong data augmentations, and achieves sample efficiency, asymptotic performance, and generalization that is competitive or better than previous state-of-the-art methods in all tasks considered, at a lower computational cost. Finally, we show that our method scales to RL with Vision Transformers (ViT) \citep{Dosovitskiy2020AnII}. We find that ViT-based architectures are especially prone to overfitting, and data augmentation may therefore be a key component for large-scale RL.

\section{Related Work}
\label{sec:related-work}
\vspace{-0.05in}
\textbf{Representation Learning.}
Learning visual invariances using data augmentation and self-supervised objectives has proven highly successful in computer vision \citep{pathak2016context, noroozi2016unsupervised, zhang2016colorful, wu2018unsupervised, oord2018representation, tian2019contrastive, Xu2020RobustAG, He2020MomentumCF, chen2020simple}. For example, Chen et al.~\citep{chen2020simple} perform an extensive study on data augmentation (e.g. random cropping and image distortions) for contrastive learning, and show that representations pre-trained with such transformations transfer effectively to downstream tasks. While our work also uses data augmentation for learning visual invariances, we leverage the $Q$-objective of deep $Q$-learning algorithms instead of auxiliary representation learning tasks.

\textbf{Visual Learning for RL.}
Numerous methods have been proposed with the goal of improving sample efficiency \citep{jaderberg2016reinforcement, Shelhamer2017LossII, oord2018representation, yarats2019improving, lin2019adaptive, srinivas2020curl, stooke2020atc, schwarzer2020data,yarats2021reinforcement} of image-based RL. Recently, using self-supervision to improve generalization in RL has also gained interest \citep{Zhang2020LearningIR, Pathak2019SelfSupervisedEV, sekar2020planning, Agarwal2021ContrastiveBS, hansen2021deployment, hansen2021softda, Wang2021UnsupervisedVA}. Notably, Zhang et al.~\citep{Zhang2020LearningIR} and Agarwal et al.~\citep{Agarwal2021ContrastiveBS} propose to learn behavioral similarity embeddings via auxiliary tasks (bisimulation metrics and contrastive learning, respectively), and Hansen et al.~\citep{hansen2021softda} learn visual invariances through an auxiliary prediction task. While these results are encouraging, it has also been shown in~\citep{jaderberg2016reinforcement, lin2019adaptive, hansen2021deployment, Yu2020GradientSF, Lyle2021OnTE} that the best choice of auxiliary tasks depends on the particular RL task, and that joint optimization with sub-optimally chosen tasks can lead to gradient interference. We achieve competitive sample-efficiency and generalization results \textit{without} the need for carefully chosen auxiliary tasks, and our method is therefore applicable to a larger variety of RL tasks.

\textbf{Data Augmentation and Randomization for RL.}
Our work is directly inspired by previous work on generalization in RL by domain randomization \citep{Tobin_2017, pinto2017asymmetric, Peng_2018, Ramos_2019, Chebotar2019ClosingTS} and data augmentation \citep{Lee2019ASR, cobbe2018quantifying, Wang2020ImprovingGI, laskin2020reinforcement, kostrikov2020image, Raileanu2020AutomaticDA, stooke2020atc, hansen2021softda}. For example, Tobin et al.~\citep{Tobin_2017} show that a neural network trained for object localization in a simulation with randomized visual augmentations improves real world generalization. Similarly, Lee et al.\citep{Lee2019ASR} show that application of a random convolutional layer to observations during training improve generalization in 3D navigation tasks. More recently, extensive studies on data augmentation~\citep{laskin2020reinforcement, kostrikov2020image} have been conducted with RL, and conclude that, while small random crops and translations can improve sample efficiency, most data augmentations \textit{decrease} sample efficiency and cause divergence. We illuminate main causes of instability, and propose a framework for data augmentation in deep $Q$-learning algorithms that drastically improves stability and generalization.

\textbf{Improving Deep $Q$-Learning.} While deep $Q$-learning algorithms such as Deep $Q$-Networks (DQN) \citep{mnih2013playing} have achieved impressive results in image-based RL, the temporal difference objective is known to have inherent instabilities when used in conjunction with function approximation and off-policy data \citep{Sutton2018}. Therefore, a variety of algorithmic improvements have been proposed to improve convergence \citep{Hasselt2016DeepRL, Wang2016DuelingNA, Hausknecht2015DeepRQ, Hasselt2016LearningVA, Schaul2016PrioritizedER, Lillicrap2016ContinuousCW, Fujimoto2018AddressingFA, Fortunato2018NoisyNF, Hessel2018RainbowCI}. For example, Hasselt et al. \citep{Hasselt2016DeepRL} reduce overestimation of $Q$-values by decomposing the target $Q$-value estimation into action selection and action evaluation using separate networks. Lillicrap et al. \citep{Lillicrap2016ContinuousCW} reduce target variance by defining the target $Q$-network as a slow-moving average of the online $Q$-network. Our method also improves $Q$-value estimation, but we specifically address the instability of deep $Q$-learning algorithms on augmented data.

\section{Preliminaries}
\label{sec:preliminaries}
\vspace{-0.025in}
\textbf{Problem formulation.}
We formulate the interaction between environment and policy as a Markov Decision Process (MDP) \citep{bellman1957mdp} $\mathcal{M}=\langle\mathcal{S}, \mathcal{A}, \mathcal{P}, r, \gamma\rangle$, where $\mathcal{S}$ is the state space, $\mathcal{A}$ is the action space, $\mathcal{P}\colon\mathcal{S}\times\mathcal{A} \mapsto \mathcal{S}$ is the state transition function that defines a conditional probability distribution $\mathcal{P}\left(\cdot | \mathbf{s}_{t}, \mathbf{a}_{t}\right)$ over all possible next states given a state $\mathbf{s}_{t} \in \mathcal{S}$ and action $\mathbf{a}_{t} \in \mathcal{A}$ taken at time $t$, $r\colon \mathcal{S}\times\mathcal{A} \mapsto \mathbb{R}$ is a reward function, and $\gamma \in [0, 1)$ is the discount factor. Because image observations only offer partial state observability \citep{kaelbling1998pomdp}, we define a state $\mathbf{s}_{t}$ as a sequence of $k+1$ consecutive frames $(\mathbf{o}_{t}, \mathbf{o}_{t-1},\dots,\mathbf{o}_{t-k}),~\mathbf{o} \in \mathcal{O}$, where $\mathcal{O}$ is the high-dimensional image space, as proposed in Mnih et al. \citep{mnih2013playing}. The goal is then to learn a policy $\pi \colon \mathcal{S} \mapsto \mathcal{A}$ that maximizes discounted return $R_{t} =\mathbb{E}_{\Gamma\sim\pi}[\sum_{t=1}^{T} \gamma^{t} r(\mathbf{s}_{t}, \mathbf{a}_{t}) ]$ along a trajectory $\Gamma = (\mathbf{s}_{0}, \mathbf{s}_{1},\dots,\mathbf{s}_{T})$ obtained by following policy $\pi$ from an initial state $\mathbf{s}_{0} \in \mathcal{S}$ to a state $\mathbf{s}_{T}$ with state transitions sampled from $\mathcal{P}$, and $\pi$ is parameterized by a collection of learnable parameters $\theta$. For clarity, we hereon generically denote parameterization with subscript, e.g. $\pi_{\theta}$. We further aim to learn parameters $\theta$ s.t. $\pi_{\theta}$ generalizes well (i.e., obtains high discounted return) to unseen MDPs, which is generally unfeasible without further assumptions about the structure of the space of MDPs. In this work, we focus on generalization to MDPs $\overline{\mathcal{M}} = \langle \overline{\mathcal{S}}, \mathcal{A}, \mathcal{P}, r, \gamma\rangle$, where states $\overline{\mathbf{s}}_{t} \in \overline{\mathcal{S}}$ are constructed from observations $\overline{\mathbf{o}}_{t} \in \overline{\mathcal{O}},~\mathcal{O} \subseteq \overline{\mathcal{O}}$ of a \textit{perturbed} observation space $\overline{\mathcal{O}}$ (e.g. unseen visuals), and $\overline{\mathcal{M}} \sim \mathbb{M}$ for a space of MDPs $\mathbb{M}$.

\textbf{Deep $Q$-Learning.}
Common model-free off-policy RL algorithms aim to estimate an optimal state-action value function $Q^{*} \colon \mathcal{S} \times \mathcal{A} \mapsto \mathbb{R}$ as $Q_{\theta}(\mathbf{s}, \mathbf{a}) \approx Q^{*}(\mathbf{s}, \mathbf{a}) = \max_{\pi_{\theta}} \mathbb{E}\left[ R_{t} | \mathbf{s}_{t} = \mathbf{s}, \mathbf{a}_{t} = \mathbf{a} \right]$ using function approximation. In practice, this is achieved by means of the single-step Bellman residual $\left(r(\mathbf{s}_{t}, \mathbf{a}_{t}) + \gamma \max_{\mathbf{a}'_{t}} Q^{\textnormal{tgt}}_{\psi}(\mathbf{s}_{t+1}, \mathbf{a}'_{t})\right) - Q_{\theta}(\mathbf{s}_{t}, \mathbf{a}_{t})$ \citep{Sutton1988LearningTP}, where $\psi$ parameterizes a \textit{target} state-action value function $Q^{\textnormal{tgt}}$. We can choose to minimize this residual (also known as the \textit{temporal difference} error) directly wrt $\theta$ using a mean squared error loss, which gives us the objective
\begin{equation}
    \label{eq:q-objective}
    \mathcal{L}_{Q}(\theta, \psi) = \mathbb{E}_{\mathbf{s}_{t}, \mathbf{a}_{t}, \mathbf{s}_{t+1} \sim \mathcal{B}} \left[ \frac{1}{2}\left[\left(r(\mathbf{s}_{t}, \mathbf{a}_{t}) + \gamma \max_{\mathbf{a}'_{t}} Q^{\textnormal{tgt}}_{\psi}(\mathbf{s}_{t+1}, \mathbf{a}'_{t})\right) - Q_{\theta}(\mathbf{s}_{t}, \mathbf{a}_{t}) \right]^{2} \right]\,,
\end{equation}
where $\mathcal{B}$ is a replay buffer with transitions collected by a behavioral policy \citep{Lin2004SelfimprovingRA}. From here, we can derive a greedy policy directly by selecting actions $\mathbf{a}_{t} = \arg\max_{\mathbf{a}_{t}} Q_{\theta}(\mathbf{s}_{t}, \mathbf{a}_{t})$. While $Q^{\textnormal{tgt}} = Q$ and periodically setting $\psi\longleftarrow\theta$ exactly recovers the objective of DQN \citep{mnih2013playing}, several improvements have been proposed to improve stability of Eq. \ref{eq:q-objective}, such as Double Q-learning \citep{Hasselt2016DeepRL}, Dueling $Q$-networks \citep{Wang2016DuelingNA}, updating target parameters using a slow-moving average of the online $Q$-network \citep{Lillicrap2016ContinuousCW}:
\begin{equation}
    \label{eq:critic-target-ema}
    \psi_{n+1} \longleftarrow (1 - \zeta) \psi_{n} + \zeta \theta_{n}
\end{equation}
for an iteration step $n$ and a momentum coefficient $\zeta \in (0,1]$, and others \citep{Hausknecht2015DeepRQ, Hasselt2016LearningVA, Schaul2016PrioritizedER, Fortunato2018NoisyNF, Hessel2018RainbowCI}. As computing $\max_{\mathbf{a}'_{t}} Q^{\textnormal{tgt}}_{\psi}(\mathbf{s}_{t+1}, \mathbf{a}'_{t})$ in Eq. \ref{eq:q-objective} is intractable for large and continuous action spaces, a number of prominent \textit{actor-critic} algorithms that additionally learn a policy $\pi_{\theta}(\mathbf{s}_{t}) \approx \arg\max_{\mathbf{a}_{t}} Q_{\theta}(\mathbf{s}_{t}, \mathbf{a}_{t})$ have therefore been proposed \citep{Lillicrap2016ContinuousCW, Fujimoto2018AddressingFA, haarnoja2018soft}.

\textbf{Soft Actor-Critic} (SAC) \citep{haarnoja2018soft} is an off-policy actor-critic algorithm that learns a state-action value function $Q_{\theta}$ and a stochastic policy $\pi_{\theta}$ (and optionally a temperature parameter), where $Q_{\theta}$ is optimized using a variant of the objective in Eq. \ref{eq:q-objective} and $\pi_{\theta}$ is optimized using a $\gamma$-discounted maximum-entropy objective \citep{ziebart2008maximum}. To improve stability, SAC is also commonly implemented using Double Q-learning and the slow-moving target parameters from Eq. \ref{eq:critic-target-ema}. We will in the remainder of this work describe our method in the context of a generic off-policy RL algorithm that learns a parameterized state-action value function $Q_{\theta}$, while we in our experiments discussed in Section \ref{sec:experiments} evaluate of our method using SAC as base algorithm.

\section{Pitfalls of Data Augmentation in Deep $Q$-Learning}
\label{sec:pitfalls-data-aug}
\begin{wrapfigure}[29]{r}{0.53\textwidth}%
    \centering
    \vspace{-3.5ex}
    \includegraphics[width=0.53\textwidth]{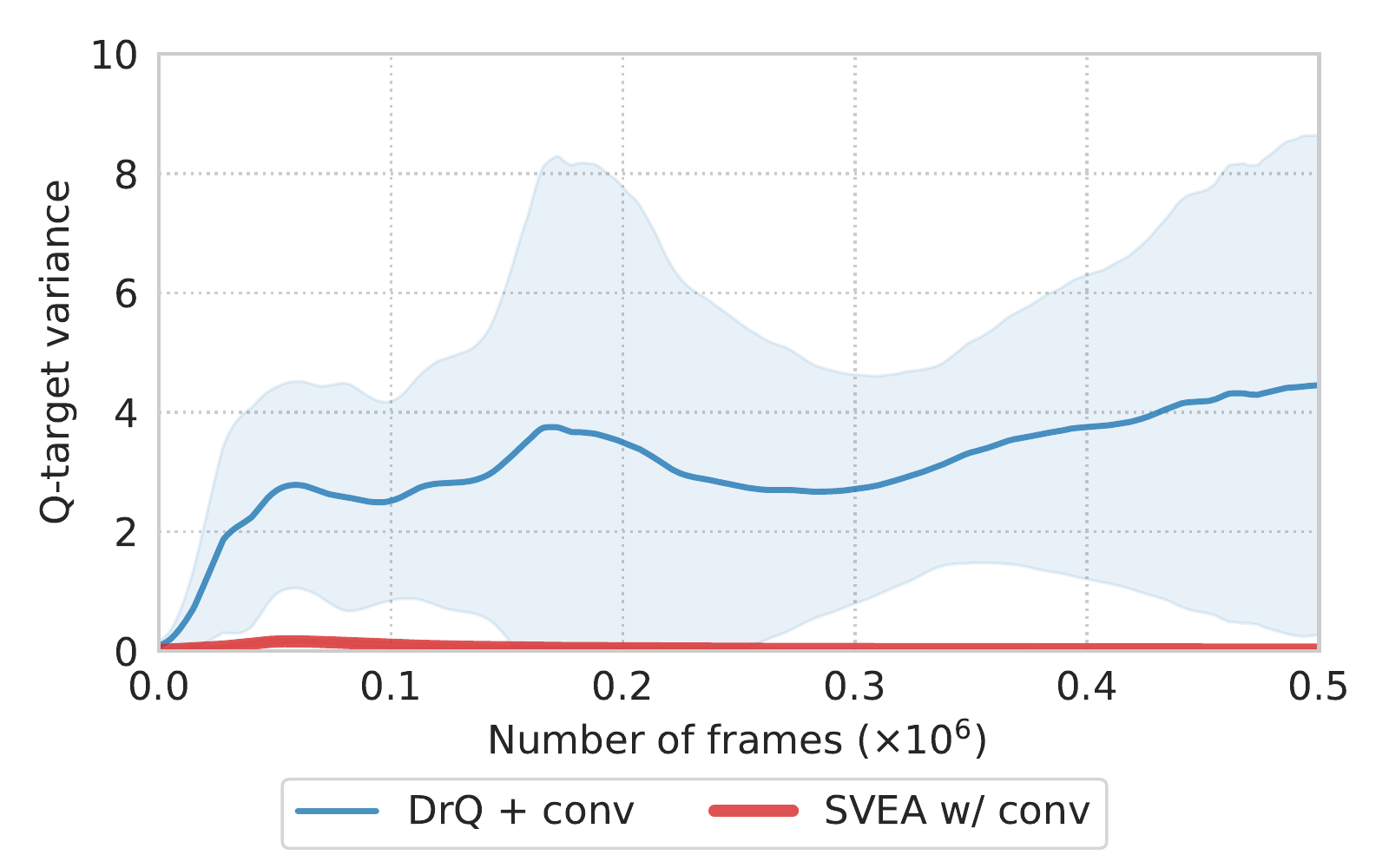}
    \includegraphics[width=0.53\textwidth]{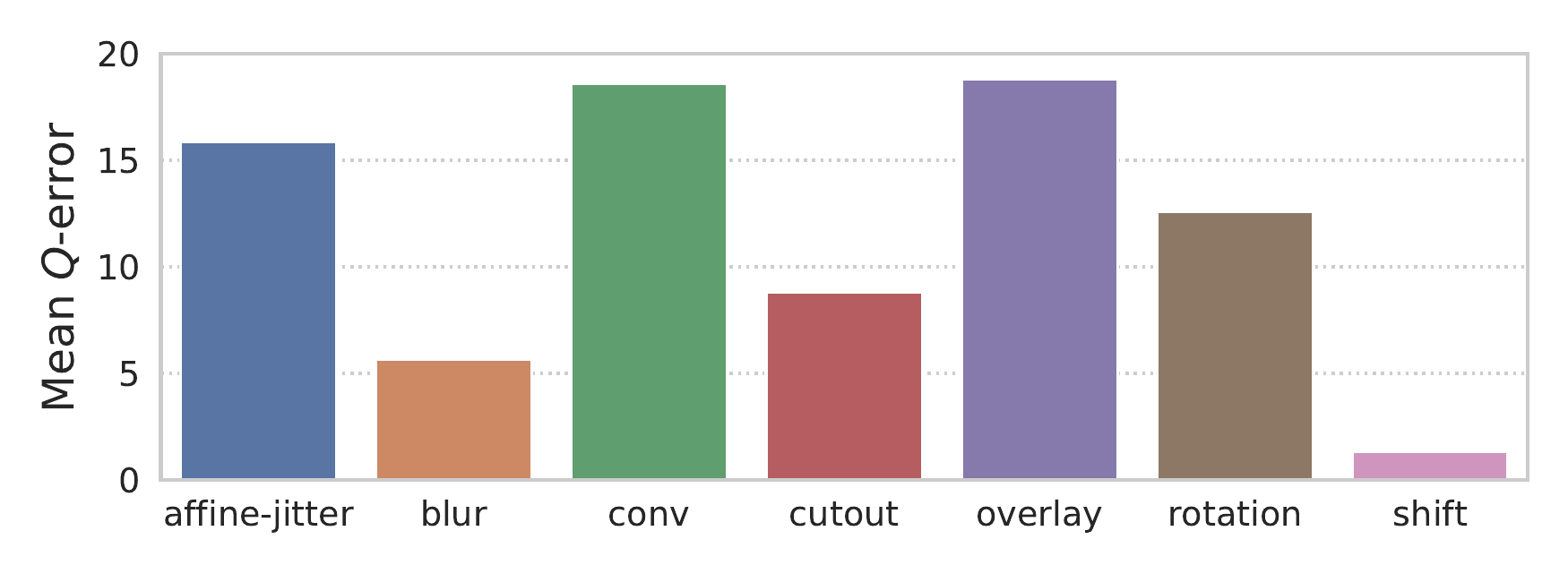}
    \vspace{-0.25in}
    \caption{\textit{(Top)} \textbf{Mean $Q$-target variance} of DrQ \protect{\citep{kostrikov2020image}} and SVEA (ours), both trained with \textit{conv} augmentation \cite{Lee2019ASR}. \textit{(Bottom)} \textbf{Mean difference in $Q$-value estimation on augmented vs. non-augmented data.} We measure mean absolute error in $Q$-value estimation from converged DrQ agents (trained with \textit{shift} augmentation) on the same observations before and after augmentation. Both figures are averages across 5 seeds for each of the 5 tasks from DMControl-GB.}
    \label{fig:dmc-q-errors}
\end{wrapfigure}

\vspace{-0.05in}
In this section, we aim to illuminate the main causes of instability from naïve application of data augmentation in $Q$-value estimation. Our goal is to learn a $Q$-function $Q_{\theta}$ for an MDP $\mathcal{M}$ that generalizes to novel MDPs $\overline{\mathcal{M}} \sim \mathbb{M}$ with unseen visuals, and we leverage data augmentation as an optimality-invariant state transformation $\tau$ to induce a bisimulation relation \citep{LarsenBisimulation1989, Givan2003EquivalenceNA} between a state $\mathbf{s}$ and its transformed (augmented) counterpart $\mathbf{s}^{\textnormal{aug}} = \tau(\mathbf{s}, \nu)$ with parameters $\nu \sim \mathcal{V}$.
\begin{definition}[Optimality-Invariant State Transformation \citep{kostrikov2020image}]
\label{def:optimality-invariant-state-transformation}
\textit{Given an MDP $\mathcal{M}$, a state transformation $\tau\colon \mathcal{S} \times \mathcal{V} \mapsto \mathcal{S}$ is an optimality-invariant state transformation if $Q(\mathbf{s}, \mathbf{a}) = Q(\tau(\mathbf{s}, \nu), \mathbf{a})~~\forall~ \mathbf{s} \in \mathcal{S},~\mathbf{a} \in \mathcal{A},~\nu \in \mathcal{V}$, where $\nu \in \mathcal{V}$ parameterizes the transformation $\tau$.}
\end{definition}
Following our definitions of $\mathcal{M}, \overline{\mathcal{M}}$ from Section \ref{sec:preliminaries}, we can further extend the concept of optimality-invariant transformations to MDPs, noting that a change of state space (e.g. perturbed visuals) itself can be described as a transformation $\overline{\tau} \colon \mathcal{S} \times \overline{\mathcal{V}} \mapsto \overline{\mathcal{S}}$ with unknown parameters $\overline{\nu} \in \overline{\mathcal{V}}$. If we choose the set of parameters $\mathcal{V}$ of a state transformation $\tau$ to be sufficiently large such that it intersects with $\overline{\mathcal{V}}$ with high probability, we can therefore expect to improve generalization to state and observation spaces not seen during training. However, while naïve application of data augmentation as in previous work \citep{laskin2020reinforcement, kostrikov2020image, stooke2020atc, schwarzer2020data} may potentially improve generalization, it can be harmful to $Q$-value estimation. We hypothesize that this is primarily because it dramatically increases the size of the observed state space, and consequently also increases variance $\operatorname{Var}\left[ Q(\tau(\mathbf{s}, \nu)) \right] \geq \operatorname{Var}\left[ Q(\mathbf{s}) \right],~\nu \sim \mathcal{V}$ when $\mathcal{V}$ is large. Concretely, we identify the following two issues:

\textbf{Pitfall 1: Non-deterministic $Q$-target.} For deep $Q$-learning algorithms, previous work \citep{laskin2020reinforcement, kostrikov2020image, stooke2020atc, schwarzer2020data} applies augmentation to both state $\mathbf{s}^{\textnormal{aug}}_{t} \triangleq \tau(\mathbf{s}_{t}, \nu)$ and successor state $\mathbf{s}^{\textnormal{aug}}_{t+1} \triangleq \tau(\mathbf{s}_{t+1}, \nu')$ where $\nu,\nu' \sim \mathcal{V}$. Compared with DQN \citep{mnih2013playing} that uses a deterministic (more precisely, periodically updated) $Q$-target, this practice introduces a non-deterministic $Q$-target $r(\mathbf{s}_{t}, \mathbf{a}_{t}) + \gamma \max_{\mathbf{a}'_{t}} Q^{\textnormal{tgt}}_{\psi}(\mathbf{s}^{\textnormal{aug}}_{t+1}, \mathbf{a}'_{t})$ depending on the augmentation parameters $\nu'$. As observed in the original DQN paper, high-variance target values are detrimental to $Q$-learning algorithms, and may cause divergence due to the ``deadly triad'' of function approximation, bootstrapping, and off-policy learning \citep{Sutton2018}. This motivates the work to introduce a slowly changing target network, and several other works have refined the $Q$-target update rule \citep{Lillicrap2016ContinuousCW, Fujimoto2018AddressingFA} to further reduce volatility. However, because data augmentation is inherently non-deterministic, it can greatly increase variance in $Q$-target estimation and exacerbates the issue of volatility, as shown in Figure \ref{fig:dmc-q-errors} (top). This is particularly troubling in actor-critic algorithms such as DDPG \citep{Lillicrap2016ContinuousCW} and SAC \citep{haarnoja2018soft}, where the $Q$-target is estimated from $(\mathbf{s}_{t+1}, \mathbf{a}'), ~\mathbf{a'} \sim \pi(\cdot | \mathbf{s}_{t+1})$, which introduces an additional source of error from $\pi$ that is non-negligible especially when $\mathbf{s}_{t+1}$ is augmented.

\textbf{Pitfall 2: Over-regularization.} Data augmentation was originally introduced in the supervised learning regime as a regularizer to prevent overfitting of high-capacity models. However, for RL, even learning a policy in the training environment is hard. While data augmentation may improve generalization, it greatly increases the difficulty of policy learning, i.e., optimizing $\theta$ for $Q_{\theta}$ and potentially a behavior network $\pi_{\theta}$. Particularly, when the temporal difference loss from Eq. \ref{eq:q-objective} cannot be well minimized, the large amount of augmented states dominate the gradient, which significantly impacts $Q$-value estimation of both augmented and unaugmented states. We refer to this issue as \textit{over-regularization} by data augmentation. Figure \ref{fig:dmc-q-errors} (bottom) shows the mean difference in $Q$-predictions made with augmented vs. unaugmented data in fully converged DrQ \citep{kostrikov2020image} agents trained with \textit{shift} augmentation. Augmentations such as affine-jitter, random convolution, and random overlay incur large differences in estimated $Q$-values. While such difference can be reduced by regularizing the optimization with each individual augmentation, we emphasize that even the minimal shift augmentation used throughout training incurs non-zero difference. Since $\psi$ is commonly chosen to be a moving average of $\theta$ as in Eq. \ref{eq:critic-target-ema}, such differences caused by over-regularization affect $Q_{\theta}$ and $Q^{\textnormal{tgt}}_{\psi}$ equally, and optimization may therefore still diverge depending on the choice of data augmentation. As such, there is an inherent trade-off between accurate $Q$-value estimation and generalization when using data augmentation. In the following section, we address these pitfalls.

\begin{figure}
    \centering
    \includegraphics[width=0.82\textwidth]{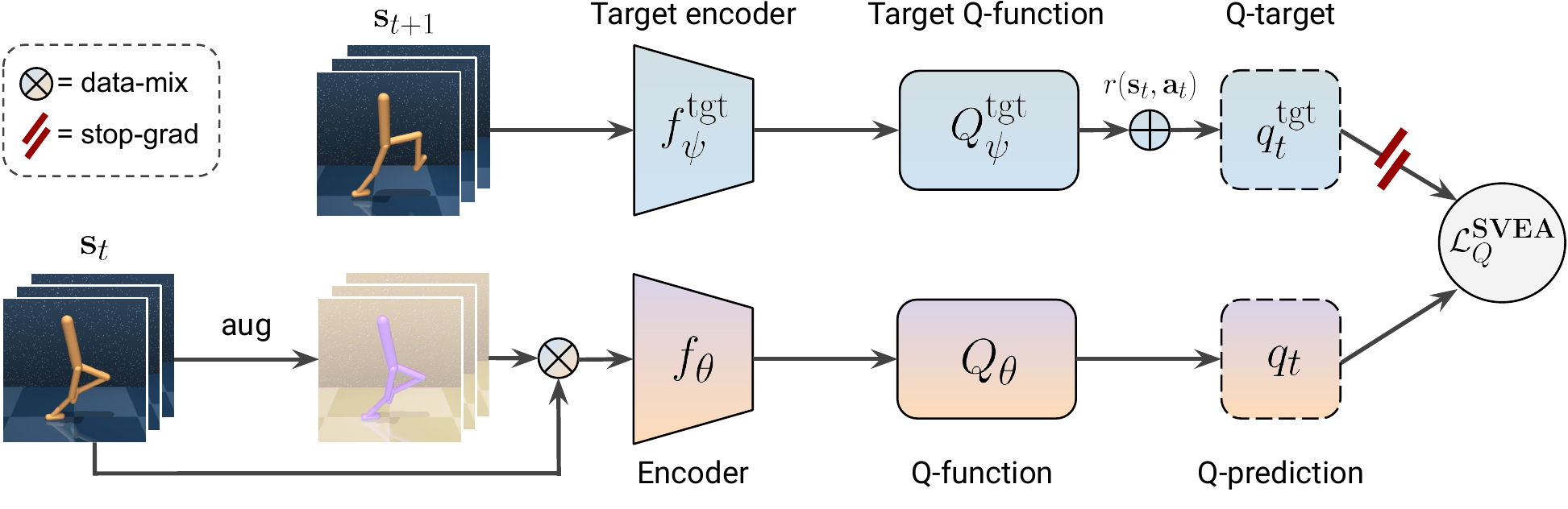}
    \vspace{-0.05in}
    \caption{\textbf{Overview.}
    An observation $\mathbf{s}_{t}$ is transformed by data augmentation $\tau(\cdot, \nu),~\nu \sim \mathcal{V}$ to produce a view $\mathbf{s}^{\textnormal{aug}}_{t}$. The $Q$-function $Q_{\theta}$ is then jointly optimized on both augmented and unaugmented data wrt the objective in Eq. \ref{eq:new-critic-loss-single-forward-pass}, with the $Q$-target of the Bellman equation computed from an unaugmented observation $\mathbf{s}_{t+1}$. We illustrate our data-mixing strategy by the $\otimes$ operator.
    }
    \label{fig:method}
    \vspace{-0.05in}
\end{figure}

\section{Method}
\label{sec:method}
\vspace{-0.125in}
We propose \textbf{SVEA}: \textbf{S}tabilized $Q$-\textbf{V}alue \textbf{E}stimation under \textbf{A}ugmentation, a general framework for visual generalization in RL by use of data augmentation. SVEA applies data augmentation in a novel learning framework leveraging two data streams -- with and without augmented data, respectively. Our method is compatible with any standard off-policy RL algorithm without changes to the underlying neural network that parameterizes the policy, and it requires no additional forward passes, auxiliary tasks, nor learnable parameters. While SVEA in principle does not make any assumptions about the structure of states $\mathbf{s}_{t} \in \mathcal{S}$, we here describe our method in the context of image-based RL.

\subsection{Architectural Overview} 
\label{sec:architectural-overview}
\vspace{-0.025in}
An overview of the SVEA architecture is provided in Figure \ref{fig:method}. Our method leverages properties of common neural network architectures used in off-policy RL without introducing additional learnable parameters. We subdivide the neural network layers and corresponding learnable parameters of a state-action value function into sub-networks $f_{\theta}$ (denoted the state \textit{encoder}) and $Q_{\theta}$ (denoted the $Q$-\textit{function}) s.t $q_{t} \triangleq Q_{\theta}(f_{\theta}(\mathbf{s}_{t}), \mathbf{a}_{t})$ is the predicted $Q$-value corresponding to a given state-action pair $(\mathbf{s}_{t}, \mathbf{a}_{t})$. We similarly define the target state-action value function s.t. $q^{\textnormal{tgt}}_{t} \triangleq r(\mathbf{s}_{t}, \mathbf{a}_{t}) + \gamma \max_{\mathbf{a}'_{t}} Q^{\textnormal{tgt}}_{\psi}(f^{^{\textnormal{tgt}}}_{\psi}(\mathbf{s}_{t+1}), \mathbf{a}')$ is the target $Q$-value for $(\mathbf{s}_{t}, \mathbf{a}_{t})$, and we define parameters $\psi$ as an exponential moving average of $\theta$ as in Eq. \ref{eq:critic-target-ema}. Depending on the choice of underlying algorithm, we may choose to additionally learn a parameterized policy $\pi_{\theta}$ that shares encoder parameters with $Q_{\theta}$ and selects actions $\mathbf{a}_{t} \sim \pi_{\theta}(\cdot|f_{\theta}(\mathbf{s}_{t}))$.

To circumvent erroneous bootstrapping from augmented data (as discussed in Section \ref{sec:pitfalls-data-aug}), we strictly apply data augmentation in $Q$-value estimation of the \textit{current} state $\mathbf{s}_{t}$, \textit{without} applying data augmentation to the successor state $\mathbf{s}_{t+1}$ used in Eq. \ref{eq:q-objective} for bootstrapping with $Q^{\textnormal{tgt}}_{\psi}$ (and $\pi_{\theta}$ if applicable), which addresses Pitfall 1. If $\pi_{\theta}$ is learned (i.e., SVEA is implemented with an actor-critic algorithm), we also optimize it strictly from unaugmented data. To mitigate over-regularization in optimization of $f_{\theta}$ and $Q_{\theta}$ (Pitfall 2), we further employ a modified $Q$-objective that leverages both augmented and unaugmented data, which we introduce in the following section.

\subsection{Learning Objective}
\label{sec:learning-objective}
\vspace{-0.05in}
Our method redefines the temporal difference objective from Eq. \ref{eq:q-objective} to better leverage data augmentation. First, recall that $q^{\textnormal{tgt}}_{t} = r(\mathbf{s}_{t}, \mathbf{a}_{t}) + \gamma \max_{\mathbf{a}'_{t}} Q^{\textnormal{tgt}}_{\psi}(f^{^{\textnormal{tgt}}}_{\psi}(\mathbf{s}_{t+1}), \mathbf{a}')$. Instead of learning to predict $q^{\textnormal{tgt}}_{t}$ only from state $\mathbf{s}_{t}$, we propose to minimize a nonnegative linear combination of $\mathcal{L}_{Q}$ over two individual data streams, $\textcolor{citecolor}{\mathbf{s}_{t}}$ and $\textcolor{Fuchsia}{\mathbf{s}^{\textnormal{aug}}_{t} = \tau(\mathbf{s}_{t}, \nu),~\nu \sim \mathcal{V}}$, which we define as the objective
\begin{align}
    \mathcal{L}^{\textbf{SVEA}}_{Q}(\theta, \psi) & \triangleq \textcolor{citecolor}{\alpha \mathcal{L}_{Q}\left(\mathbf{s}_{t}, q^{\textnormal{tgt}}_{t}; \theta, \psi\right)} + \textcolor{Fuchsia}{\beta \mathcal{L}_{Q}\left(\mathbf{s}^{\textnormal{aug}}_{t}, q^{\textnormal{tgt}}_{t}; \theta,\psi\right)} \\
    \label{eq:new-critic-loss}
    & = \mathbb{E}_{\mathbf{s}_{t}, \mathbf{a}_{t}, \mathbf{s}_{t+1} \sim \mathcal{B}} \left[ \textcolor{citecolor}{\alpha \left\| Q_{\theta}(f_{\theta}(\mathbf{s}_{t}), \mathbf{a}_{t}) - q^{\textnormal{tgt}}_{t} \right\|^{2}_{2}} + \textcolor{Fuchsia}{\beta \left\| Q_{\theta}(f_\theta(\mathbf{s}^{\textnormal{aug}}_{t}), \mathbf{a}_{t}) - q^{\textnormal{tgt}}_{t} \right\|^{2}_{2}} \right] \,,
\end{align}
where $\textcolor{citecolor}{\alpha}, \textcolor{Fuchsia}{\beta}$ are constant coefficients that balance the ratio of the \textcolor{citecolor}{unaugmented} and \textcolor{Fuchsia}{augmented} data streams, respectively, and $q^{\textnormal{tgt}}_{t}$ is computed strictly from unaugmented data. $\mathcal{L}^{\textbf{SVEA}}_{Q}(\theta, \psi)$ serves as a \textit{data-mixing} strategy that oversamples unaugmented data as an implicit variance reduction technique. As we will verify empirically in Section \ref{sec:experiments}, data-mixing is a simple and effective technique for variance reduction that works well in tandem with our proposed modifications to bootstrapping. For $\alpha=\beta$, the objective in Eq. \ref{eq:new-critic-loss} can be evaluated in a single, batched forward-pass by rewriting it as:
\begin{align}
    \mathbf{g}_{t} & = \left[ \textcolor{citecolor}{\mathbf{s}_{t}}, \textcolor{Fuchsia}{\tau(\mathbf{s}_{t}, \nu)}\right]_{\textnormal{N}}
    \\
    h_{t} & = \left[ q^{\textnormal{tgt}}_{t}, q^{\textnormal{tgt}}_{t} \right]_{\textnormal{N}} \\
    \label{eq:new-critic-loss-single-forward-pass}
     \mathcal{L}^{\textbf{SVEA}}_{Q}(\theta, \psi) & = \mathbb{E}_{\mathbf{s}_{t}, \mathbf{a}_{t}, \mathbf{s}_{t+1} \sim \mathcal{B},~\nu \sim \mathcal{V}} \left[
    (\alpha + \beta) \left\| Q_{\theta}(f_\theta(\mathbf{g}_{t}), \mathbf{a}_{t}) - h_{t} \right\|^{2}_{2} \right]\,,
\end{align}
where $\left[\cdot\right]_{\textnormal{N}}$ is a concatenation operator along the batch dimension $N$ for $\mathbf{s}_{t}, \mathbf{s}^{\textnormal{aug}}_{t} \in \mathbb{R}^{N\times C\times H\times W}$ and $q^{\textnormal{tgt}}_{t} \in \mathbb{R}^{N\times1}$, which is illustrated as $\otimes$ in Figure \ref{fig:method}. Empirically, we find $\alpha=0.5,\beta=0.5$ to be both effective and practical to implement, which we adopt in the majority of our experiments. However, more sophisticated schemes for selecting $\alpha,\beta$ and/or varying them as training progresses could be interesting directions for future research. If the base algorithm of choice learns a policy $\pi_{\theta}$, its objective $\mathcal{L}_{\pi}(\theta)$ is optimized solely on unaugmented states $\mathbf{s}_{t}$ without changes to the objective, and a \texttt{stop-grad} operation is applied after $f_{\theta}$ to prevent non-stationary gradients of $\mathcal{L}_{\pi}(\theta)$ from interfering with $Q$-value estimation, i.e., only the objective from Eq. \ref{eq:new-critic-loss} or optionally Eq. \ref{eq:new-critic-loss-single-forward-pass} updates $f_{\theta}$ using stochastic gradient descent. As described in Section \ref{sec:architectural-overview}, parameters $\psi$ are updated using an exponential moving average of $\theta$ and a \texttt{stop-grad} operation is therefore similarly applied after $Q^{\textnormal{tgt}}_{\psi}$. We summarize our method for $\alpha=\beta$ applied to a generic off-policy algorithm in Algorithm \ref{alg:pseudo-code}.
\vspace{-0.075in}
\begin{algorithm}
\caption{~~Generic \textbf{SVEA} off-policy algorithm~~({\color{BrickRed}$\blacktriangleright$~naïve augmentation},~~{\color{citecolor}$\blacktriangleright$~our modifications})}
\label{alg:pseudo-code}
\begin{algorithmic}[1]
\algnotext{EndFor}
\Statex $\theta, \theta_{\pi}, \psi$: randomly initialized network parameters, $\psi \longleftarrow \theta$ \hfill $\vartriangleright$~~Initialize $\psi$ to be equal to $\theta$
\Statex $\eta, \zeta$: learning rate and momentum coefficient
\Statex $\alpha, \beta$: loss coefficients, \textit{default:} $(\alpha=0.5, \beta=0.5)$
\For{timestep $t=1...T$}
\Statex \textbf{act:}
\State $\mathbf{a}_{t} \sim \pi_{\theta}\left(\cdot | f_{\theta}(\mathbf{s}_{t})\right)$ \hfill $\vartriangleright$~~Sample action from policy
\State $\mathbf{s}'_{t} \sim \mathcal{P}(\cdot | \mathbf{s}_{t}, \mathbf{a}_{t})$ \hfill $\vartriangleright$~~Sample transition from environment
\State $\mathcal{B} \leftarrow \mathcal{B} \cup (\mathbf{s}_{t}, \mathbf{a}_{t}, r(\mathbf{s}_{t}, \mathbf{a}_{t}), \mathbf{s}'_{t})$ \hfill $\vartriangleright$~~Add transition to replay buffer
\Statex \textbf{update:}
\State $\{\mathbf{s}_{i}, \mathbf{a}_{i}, r(\mathbf{s}_{i}, \mathbf{a}_{i}), \mathbf{s}'_{i}~|~i = 1...N \} \sim \mathcal{B}$ \hfill $\vartriangleright$~~Sample batch of transitions
\State {\color{BrickRed}$\mathbf{s}_{i} = \tau(\mathbf{s}_{i}, \nu_{i}),~\mathbf{s}'_{i} = \tau(\mathbf{s}'_{i}, \nu'_{i}),~\nu_{i},\nu'_{i} \sim \mathcal{V}$ \hfill $\blacktriangleright$~~Naïve application of data augmentation}
\For{transition $i=1..N$}
\State $\theta_{\pi} \longleftarrow \theta_{\pi} - \eta \nabla_{\theta_{\pi}} \mathcal{L}_{\pi} \left(\mathbf{s}_{i}; \theta_{\pi} \right)$~~(if applicable) \hfill $\vartriangleright$~~Optimize $\pi_{\theta}$ with SGD
\State $q^{\textnormal{tgt}}_{i} = r(\mathbf{s}_{i}, \mathbf{a}_{i}) + \gamma \max_{\mathbf{a}'_{i}} Q^{\textnormal{tgt}}_{\psi}(f^{\textnormal{tgt}}_{\psi}(\mathbf{s}'_{i}), \mathbf{a}'_{i})$ \hfill $\vartriangleright$~~Compute $Q$-target
\State \textcolor{citecolor}{$\mathbf{s}^{\textnormal{aug}}_{i} = \tau(\mathbf{s}_{i}, \nu_{i}),~\nu_{i} \sim \mathcal{V}$} \hfill {\color{citecolor}$\blacktriangleright$~~Apply stochastic data augmentation}
\State \textcolor{citecolor}{$\mathbf{g}_{i} = \left[ \mathbf{s}_{i}, \mathbf{s}^{\textnormal{aug}}_{i}\right]_{\textnormal{N}},~h_{i} = \left[ q^{\textnormal{tgt}}_{i}, q^{\textnormal{tgt}}_{i} \right]_{\textnormal{N}}$} \hfill {\color{citecolor}$\blacktriangleright$~~Pack data streams}
\State \textcolor{citecolor}{$\theta \longleftarrow \theta - \eta \nabla_{\theta}  \mathcal{L}^{\textbf{SVEA}}_{Q} \left(\mathbf{g}_{i}, h_{i}; \theta, \psi \right)$} \hfill {\color{citecolor}$\blacktriangleright$~~Optimize $f_{\theta}$ and $Q_{\theta}$ with SGD}
\State $\psi \longleftarrow (1 - \zeta) \psi + \zeta \theta$ \hfill $\vartriangleright$~~Update $\psi$ using EMA of $\theta$
\EndFor
\EndFor
\end{algorithmic}
\end{algorithm}

\begin{wrapfigure}[9]{r}{0.5\textwidth}%
    \vspace{-6ex}
    \centering
    \includegraphics[width=0.5\textwidth]{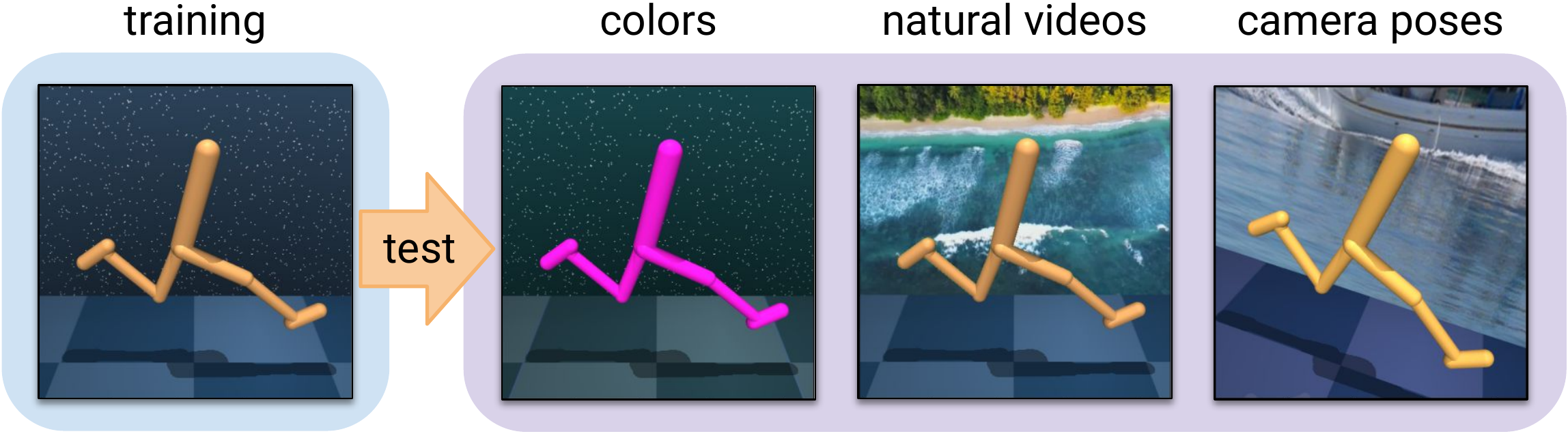}
    \caption{\textbf{Experimental setup.} Agents are trained in a fixed environment and are expected to generalize to novel environments with e.g. random colors, backgrounds, and camera poses.}
    \label{fig:dmc-test-envs}
\end{wrapfigure}

\vspace{-0.125in}
\section{Experiments}
\label{sec:experiments}
\vspace{-0.075in}
We evaluate both sample efficiency, asymptotic performance, and generalization of our method and a set of strong baselines using both ConvNets and Vision Transformers (ViT) \citep{Dosovitskiy2020AnII} in tasks from DeepMind Control Suite (DMControl) \citep{deepmindcontrolsuite2018} as well as a set of robotic manipulation tasks. DMControl offers challenging and diverse continuous control tasks and is widely used as a benchmark for image-based RL \citep{Hafner2019LearningLD, Hafner2020DreamTC, yarats2019improving, srinivas2020curl, laskin2020reinforcement, kostrikov2020image}. To evaluate generalization of our method and baselines, we test methods under challenging distribution shifts (as illustrated in Figure \ref{fig:dmc-test-envs}) from the DMControl Generalization Benchmark (DMControl-GB) \citep{hansen2021softda}, the Distracting Control Suite (DistractingCS) \citep{Stone2021TheDC}, as well as distribution shifts unique to the robotic manipulation environment. Code is available at \url{https://github.com/nicklashansen/dmcontrol-generalization-benchmark}.

\textbf{Setup.} We implement our method and baselines using SAC \citep{haarnoja2018soft} as base algorithm, and we apply random shift augmentation to all methods by default. This makes our base algorithm equivalent to DrQ \citep{kostrikov2020image} when K=1,M=1; we refer to the base algorithm as \textit{unaugmented} and consider stability under additional data augmentation. We use the \textbf{same} network architecture and hyperparameters for \textbf{all} methods (whenever applicable), and adopt the setup from \citet{hansen2021softda}. Observations are stacks of 3 RGB frames of size $84\times84\times3$ (and $96\times96\times3$ in ViT experiments). In the DMControl-GB and DistractingCS benchmarks, all methods are trained for 500k frames and evaluated on all 5 tasks from DMControl-GB used in prior work, and we adopt the same experimental setup for robotic manipulation. See Appendix \ref{sec:implementation-details} for hyperparameters and further details on our experimental setup.

\begin{figure}
    \includegraphics[width=\textwidth]{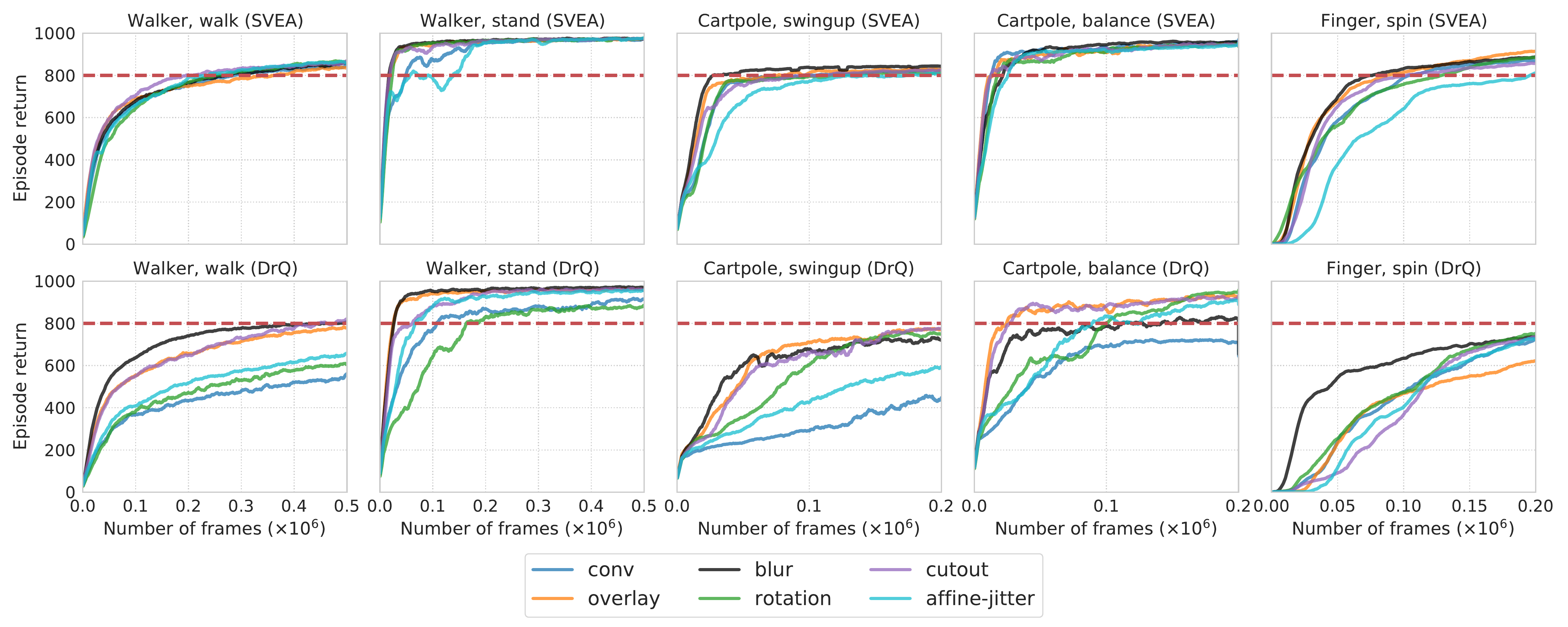}
    \vspace{-0.2in}
    \caption{\textbf{Data augmentations.} Training performance of SVEA (top) and DrQ (bottom) under 6 common data augmentations. Mean of 5 seeds. Red line at $800$ return is for visual guidance only. We omit visualization of std. deviations for clarity, but provide per-augmentation comparisons to DrQ (including std. deviations) across all tasks in Appendix \ref{sec:appendix-stability-augs}, and test performances in Appendix \ref{sec:data-augmentation}.}
    \label{fig:dmc-augs}
    \vspace{-0.125in}
\end{figure}

\begin{figure}
    \centering
    \includegraphics[width=\textwidth]{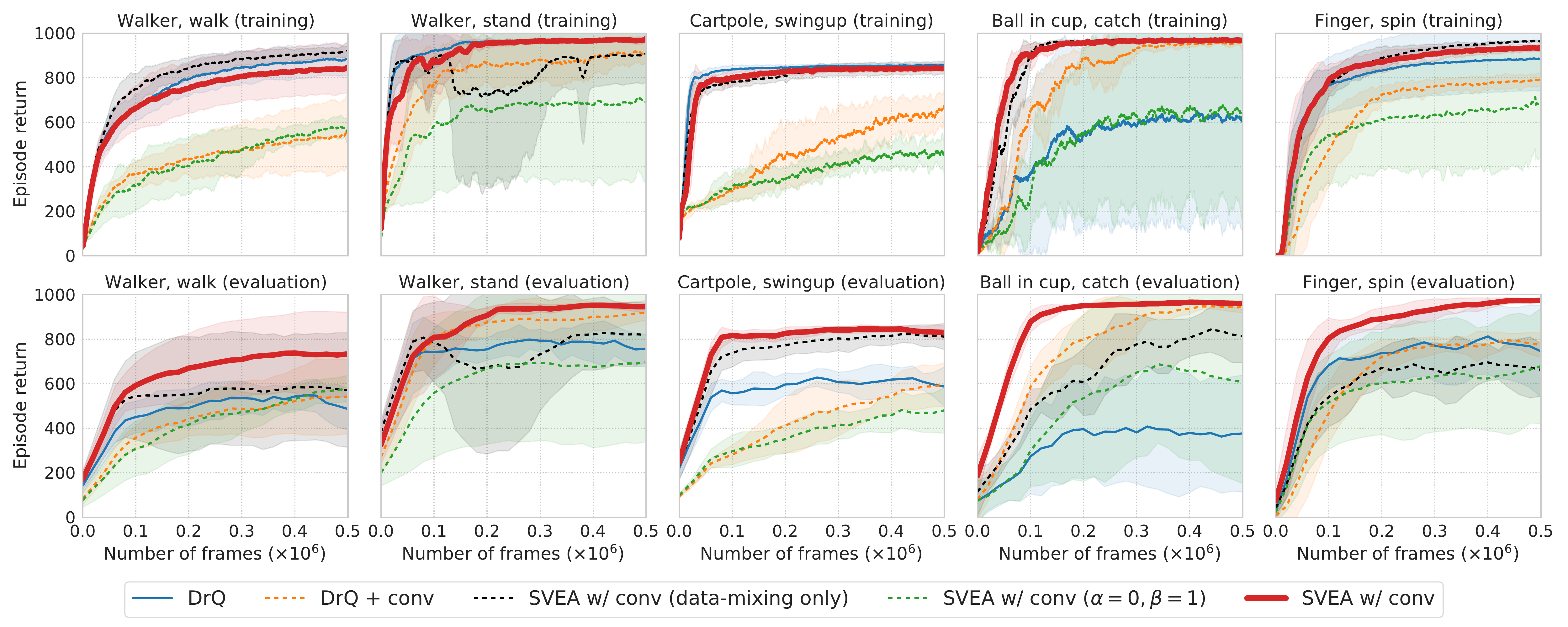}
    \vspace{-0.2in}
    \caption{\textbf{Training and test performance.} We compare SVEA to DrQ with and without random convolution augmentation, as well as a set of ablations. \textit{Data-mixing only} indiscriminately applies our data-mixing strategy to all data streams, and $(\alpha=0, \beta=1)$ only augments $Q$-predictions but without data-mixing. We find both components to contribute to SVEA's success. \textit{Top:} episode return on the training environment during training. \textit{Bottom:} generalization measured by episode return on the \texttt{color\_hard} benchmark of DMControl-GB. Mean of 5 seeds, shaded area is $\pm1$ std. deviation.}
    \label{fig:dmc-conv}
    \vspace{-0.125in}
\end{figure}

\textbf{Baselines and data augmentations.} We benchmark our method against the following strong baselines: (1) \textbf{CURL} \citep{srinivas2020curl}, a contrastive learning method for RL; (2) \textbf{RAD} that applies a random crop; (3) \textbf{DrQ} that applies a random shift; (4) \textbf{PAD} \citep{hansen2021deployment} that adapts to test environments using self-supervision; (5) \textbf{SODA} \citep{hansen2021softda} that applies data augmentation in auxiliary learning; as well as a number of ablations. We compare to the K=1,M=1 setting of DrQ by default, but also provide comparison to varying $K,M$. We experiment with a diverse set of data augmentations proposed in previous work on RL and computer vision, namely random \textit{shift} \citep{kostrikov2020image}, random convolution (denoted \textit{conv}) \citep{Lee2019ASR}, random \textit{overlay} \citep{hansen2021softda}, random \textit{cutout} \citep{cobbe2018quantifying}, Gaussian \textit{blur}, random \textit{affine-jitter}, and random \textit{rotation} \citep{laskin2020reinforcement, gidaris2018unsupervised}. We provide samples for all data augmentations in Appendix \ref{sec:data-augmentation} and test environments in Appendix \ref{sec:test-envs}.

\vspace{-0.1in}
\subsection{Stability and Generalization on DMControl}
\label{sec:experiments-dmc}
\vspace{-0.075in}

\input{tables/dmc-gb}

\begin{figure}
    \vspace{-0.025in}
    \centering
    \includegraphics[width=0.605\textwidth]{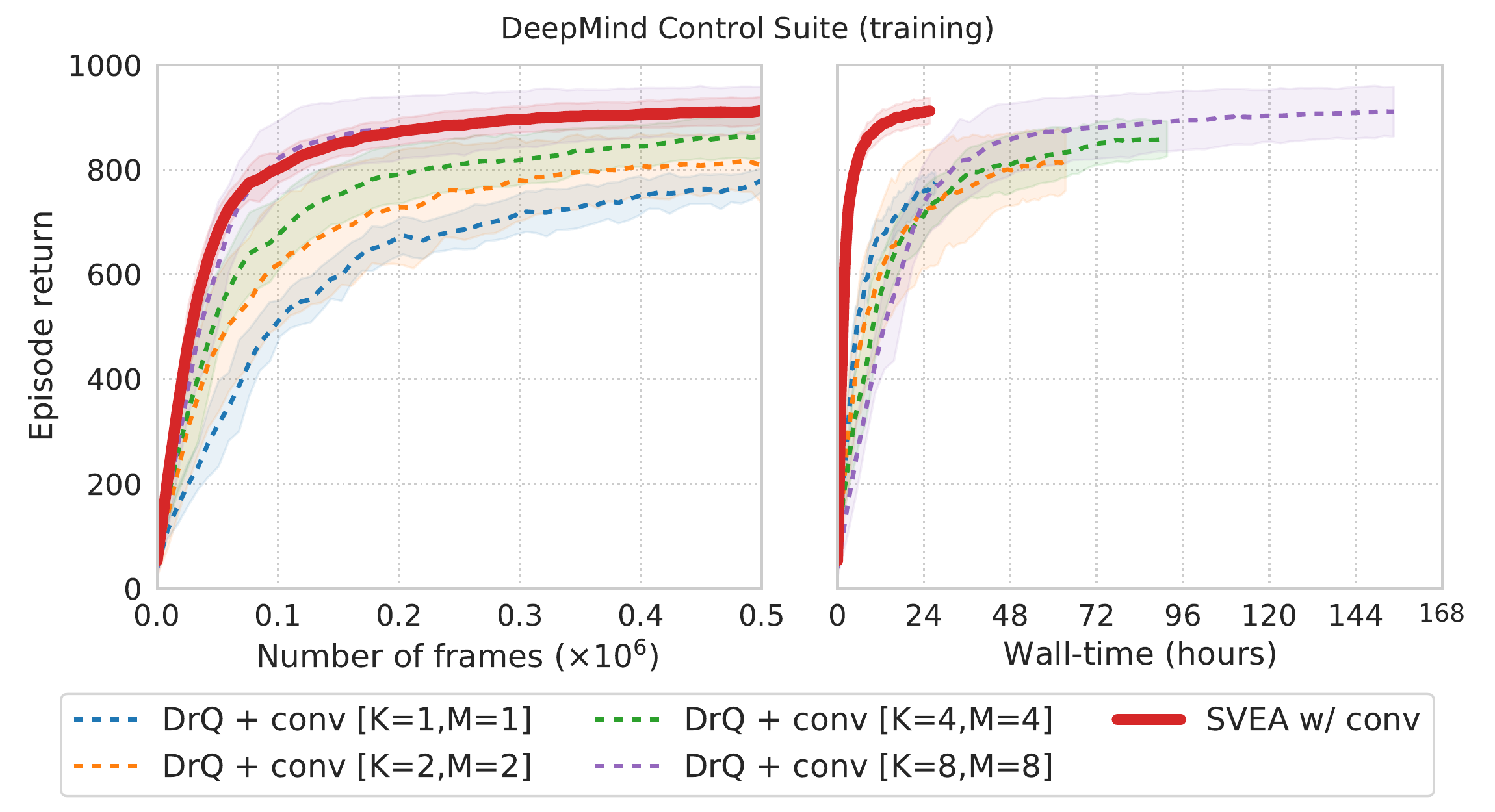}
    \includegraphics[width=0.33\textwidth]{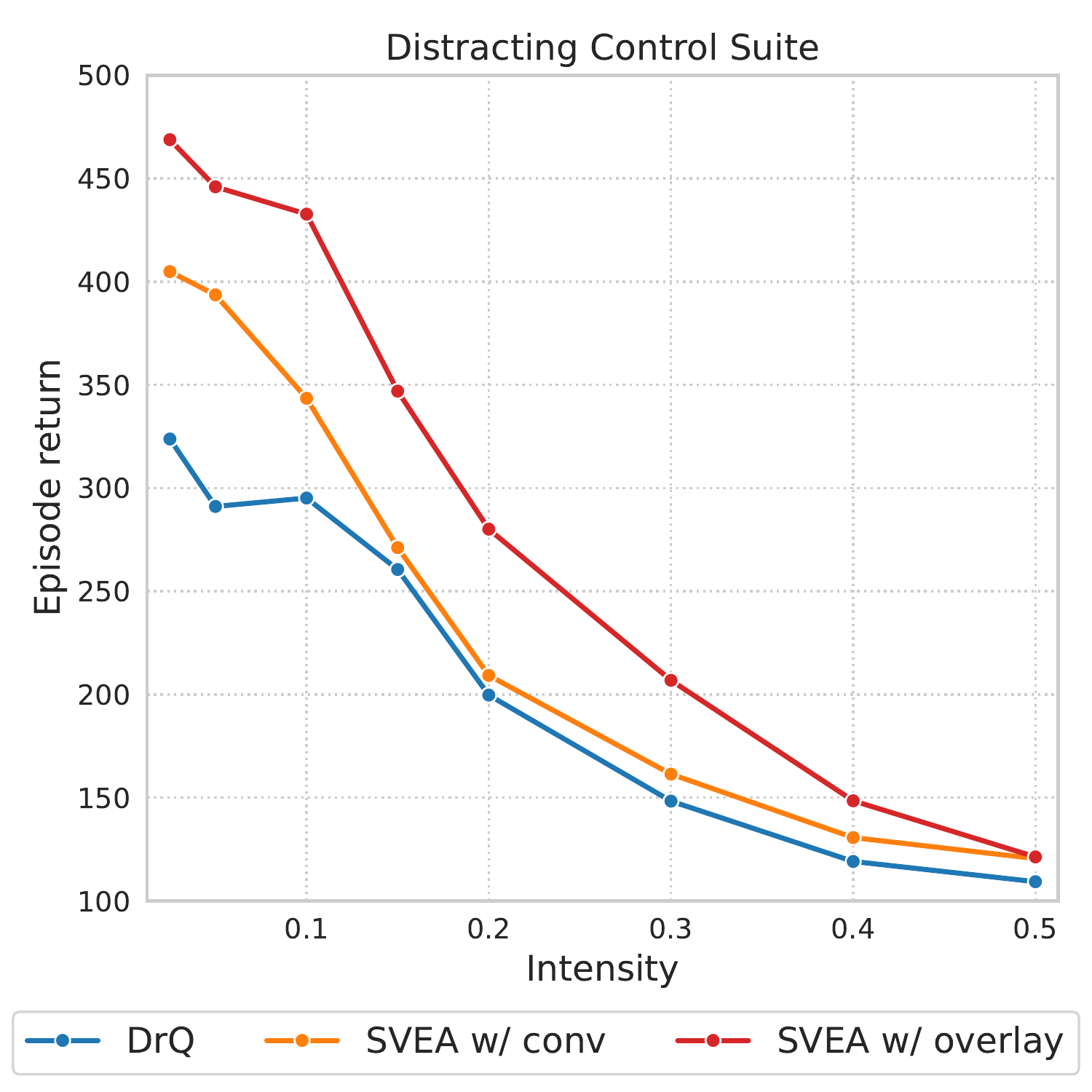}
    \vspace{-0.05in}
    \caption{\textit{(Left)} \textbf{Comparison with additional DrQ baselines.} We compare SVEA implemented with DrQ [K=1,M=1] as base algorithm to DrQ with varying values of its $K,M$ hyperparameters. All methods use the \textit{conv} augmentation (in addition to \textit{shift} augmentation used by DrQ). Results are averaged over 5 seeds for each of the 5 tasks from DMControl-GB \cite{hansen2021softda} and shaded area is $\pm1$ std. deviation across seeds. Increasing values of $K,M$ improve sample efficiency of DrQ, but at a high computational cost; DrQ uses approx. 6x wall-time to match the sample efficiency of SVEA. \textit{(Right)} \textbf{DistractingCS.} Episode return as a function of randomization intensity at test-time, aggregated across 5 seeds for each of the 5 tasks from DMControl-GB. See Appendix \ref{sec:test-envs} for per-task comparison.}
    \label{fig:drq-walltime}
    \vspace{-0.1in}
\end{figure}

\textbf{Stability.} We evaluate the stability of SVEA and DrQ under 6 common data augmentations; results are shown in Figure \ref{fig:dmc-augs}. While the sample efficiency of DrQ degrades substantially for most augmentations, SVEA is relatively unaffected by the choice of data augmentation and improves sample efficiency in $\mathbf{27}$ out of $\mathbf{30}$ instances. While the sample efficiency of DrQ can be improved by increasing its K,M parameters, we find that DrQ requires approx. 6x wall-time to match the sample efficiency of SVEA; see Figure \ref{fig:drq-walltime} \textit{(left)}. We further ablate each component of SVEA and report both training and test curves in Figure \ref{fig:dmc-conv}; we find that both components are key to SVEA's success. Because we empirically find the \textit{conv} augmentation to be particularly difficult to optimize, we provide additional stability experiments in Section \ref{sec:experiments-vit} and \ref{sec:experiments-robot} using this augmentation. See Appendix \ref{sec:ablations} for additional ablations.

\textbf{Generalization.} We compare the test performance of SVEA to 5 recent state-of-the-art methods for image-based RL on the \texttt{color\_hard} and \texttt{video\_easy} benchmarks from DMControl-GB (results in Table \ref{tab:dmc-all}), as well as the extremely challenging DistractingCS benchmark, where camera pose, background, and colors are continually changing throughout an episode (results in Figure \ref{fig:drq-walltime} (\textit{right})). We here use \textit{conv} and \textit{overlay} augmentations for fair comparison to SODA, and we report additional results on the \texttt{video\_hard} benchmark in Appendix \ref{sec:dmc-gb-additional-results}. SVEA outperforms all methods considered in $\mathbf{12}$ out of $\mathbf{15}$ instances on DMControl-GB, and at a lower computational cost than CURL, PAD, and SODA that all learn auxiliary tasks. On DistractingCS, we observe that SVEA improves generalization by $\mathbf{42\%}$ at low intensity, and its generalization degrades significantly slower than DrQ for high intensities. While generalization depends on the particular choice of data augmentation and test environments, this is an encouraging result considering that SVEA enables efficient policy learning with stronger augmentations than previous methods.

\begin{wrapfigure}[41]{r}{0.55\textwidth}%
    \vspace{-6ex}
    \centering
    \includegraphics[width=0.5\textwidth]{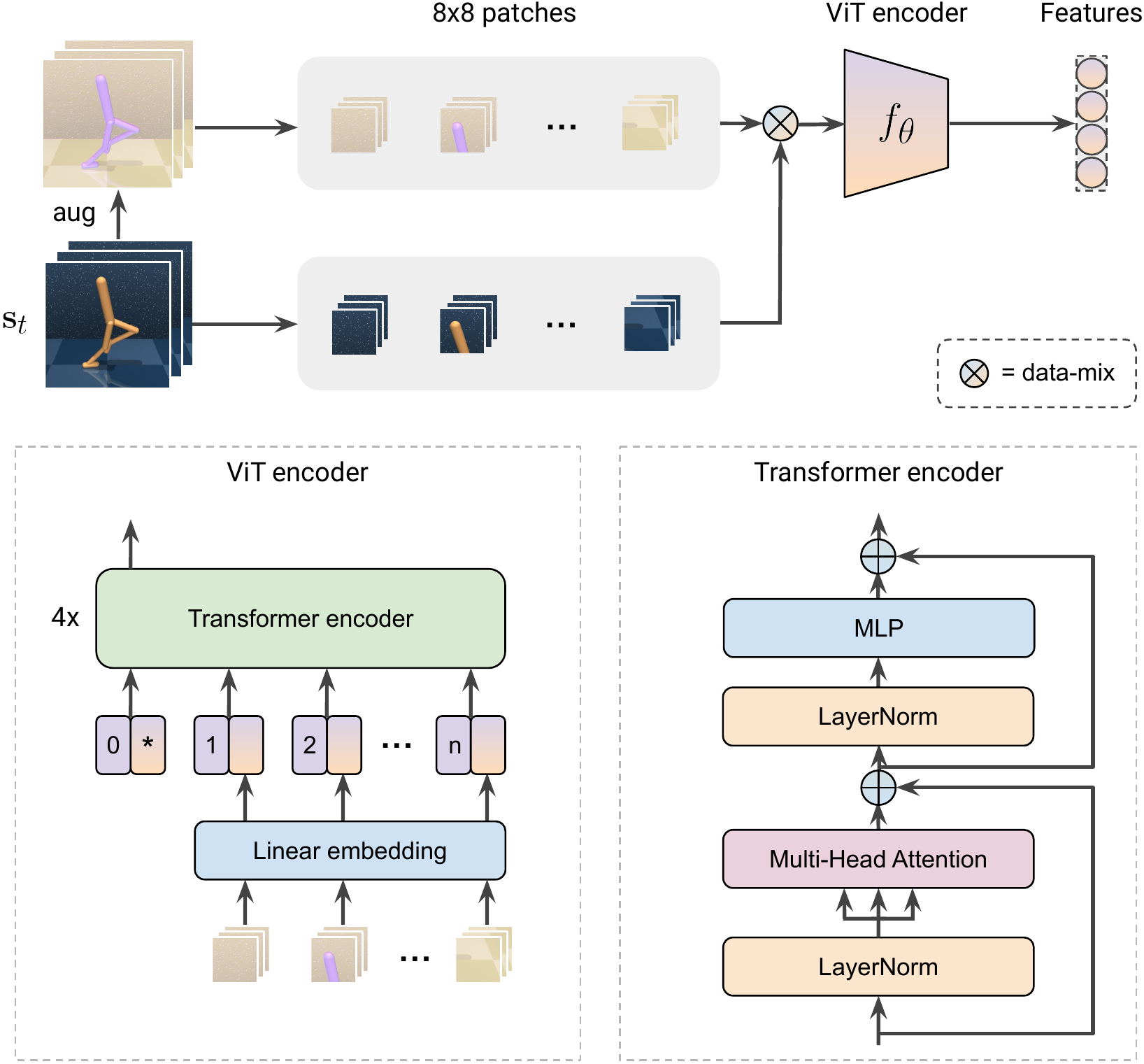}\\
    \vspace{0.05in}
    \includegraphics[width=0.55\textwidth]{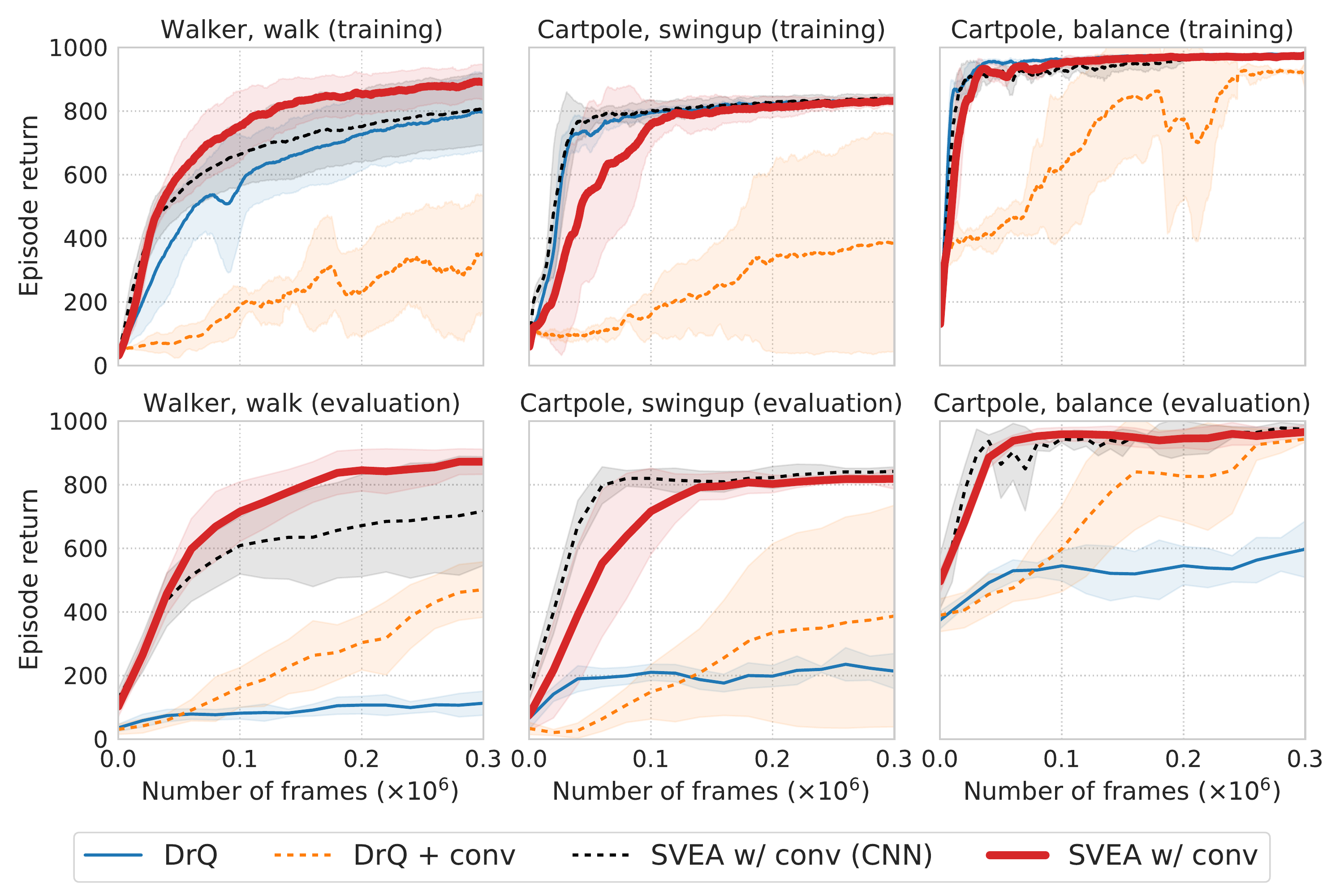}
    \caption{\textit{(Top)} \textbf{ViT architecture.} Observations are divided into $144$ non-overlapping space-time patches and linearly projected into tokens. Each token uses a learned positional encoding and we also use a learnable \texttt{class} token as in \protect{\citep{Dosovitskiy2020AnII}}. The ViT encoder consists of 4 stacked Transformer encoders \protect{\citep{Vaswani2017AttentionIA}}. \textit{(Bottom)} \textbf{RL with a ViT encoder.} Training and test performance of SVEA and DrQ using ViT encoders. We report results for three tasks and test performance is evaluated on the \texttt{color\_hard} benchmark of DMControl-GB. Mean of 5 seeds, shaded area is $\pm1$ std. deviation.}
    \label{fig:vit-architecture}
\end{wrapfigure}

\vspace{-0.05in}
\subsection{RL with Vision Transformers}
\label{sec:experiments-vit}
\vspace{-0.05in}
Vision Transformers (ViT) \citep{Dosovitskiy2020AnII} have recently achieved impressive results on downstream tasks in computer vision. We replace all convolutional layers from the previous experiments with a 4-layer ViT encoder that operates on raw pixels in $8\times8$ space-time patches, and evaluate our method using data augmentation in conjunction with ViT encoders. Importantly, we design the ViT architecture such that it roughly matches our CNN encoder in terms of learnable parameters. The ViT encoder is trained from scratch using RL, and we use the same experimental setup as in our ConvNet experiments. In particular, it is worth emphasizing that both our ViT and CNN encoders are trained using Adam \cite{Kingma2015AdamAM} as optimizer and without weight decay. See Figure \ref{fig:vit-architecture} \textit{(top)} for an architectural overview, and refer to Appendix \ref{sec:implementation-details} for additional implementation details.

Our training and test results are shown in Figure \ref{fig:vit-architecture} \textit{(bottom)}. We are, to the best of our knowledge, the first to successfully solve image-based RL tasks without CNNs. We observe that DrQ overfits significantly to the training environment compared to its CNN counterpart ($\mathbf{94}$ test return on \texttt{color\_hard} for DrQ with ViT vs. $\mathbf{569}$ with a ConvNet on the \textit{Walker, walk} task). SVEA achieves comparable sample efficiency and improves generalization by $\mathbf{706\%}$ and $\mathbf{233\%}$ on \textit{Walker, walk} and \textit{Cartpole, swingup}, respectively, over DrQ, while DrQ + conv remains unstable. Interestingly, we observe that our ViT-based implementation of SVEA achieves a mean episode return of $\mathbf{877}$ on the \texttt{color\_hard} benchmark of the challenging \textit{Walker, walk} task (vs. $\mathbf{760}$ using CNNs). SVEA might therefore be a promising technique for future research on RL with CNN-free architectures, where data augmentation appears to be especially important for generalization. We provide additional experiments with ViT encoders in Section \ref{sec:experiments-robot} and make further comparison to ConvNet encoders in Appendix \ref{sec:ablations}.

\newpage

\begin{table}[t]
    \centering
    \caption{\textbf{Generalization in robotic manipulation.} Task success rates of SVEA and DrQ with CNN and ViT encoders in the training environment, as well as aggregated success rates across 25 different test environments with randomized camera pose, colors, lighting, and background. Mean of 5 seeds.}
    \label{tab:robot}
    \vspace{0.05in}
    \resizebox{0.75\textwidth}{!}{%
    \begin{tabular}{lc|cc|cc|cc}
    \toprule
    Robotic         & Arch. & \texttt{reach} & \texttt{reach} & \texttt{mv.tgt.} & \texttt{mv.tgt.}& \texttt{push}   & \texttt{push} \\
    manipulation    & (encoder) & (train) & (test) & (train) & (test) & (train) & (test) \\\midrule
    DrQ            & CNN & $\mathbf{1.00}$ & $0.60$ & $\mathbf{1.00}$ & $0.69$ & $\mathbf{0.76}$ & $0.26$ \\
    DrQ + conv     & CNN & $0.59$ & $0.77$ & $0.60$ & $0.89$ & $0.13$ & $0.12$ \\
    \textbf{SVEA} w/ conv   & CNN & $\mathbf{1.00}$ & $\mathbf{0.89}$ & $\mathbf{1.00}$ & $\mathbf{0.96}$ & $0.72$ & $\mathbf{0.48}$ \\
    \midrule
    DrQ            & ViT & $0.93$ & $0.14$ & $\mathbf{1.00}$ & $0.16$ & $0.73$ & $0.05$ \\
    DrQ + conv     & ViT & $0.26$ & $0.67$ & $0.48$ & $\mathbf{0.82}$ & $0.08$ & $0.07$ \\
    \textbf{SVEA} w/ conv    & ViT & $\mathbf{0.98}$ & $\mathbf{0.71}$ & $\mathbf{1.00}$ & $0.81$ & $\mathbf{0.82}$ & $\mathbf{0.17}$ \\
    \bottomrule
    \end{tabular}
    }
    \vspace{-0.15in}
\end{table}

\begin{wrapfigure}[35]{r}{0.6\textwidth}%
    \vspace{-6.25ex}
    \centering
    \includegraphics[width=0.54\textwidth]{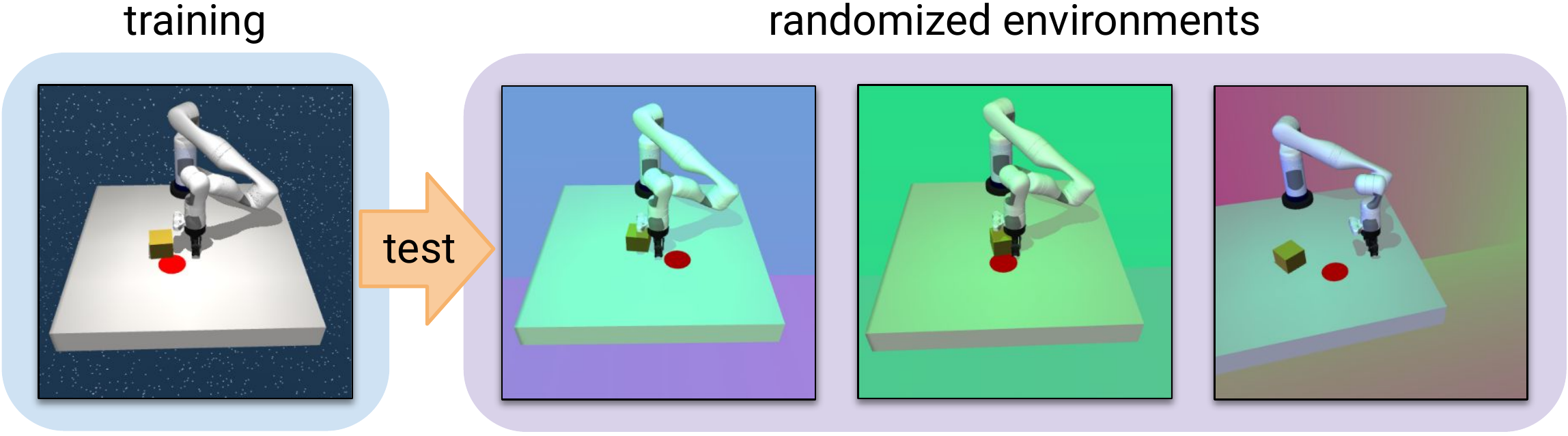}
    \vspace{-0.05in}
    \captionof{figure}{\textbf{Robotic manipulation.} Agents are trained in a fixed environment and evaluated on challenging environments with randomized colors, lighting, background, and camera pose.}
    \label{fig:robot-experimental-setup}
    \vspace{0.05in}
    \includegraphics[width=0.6\textwidth]{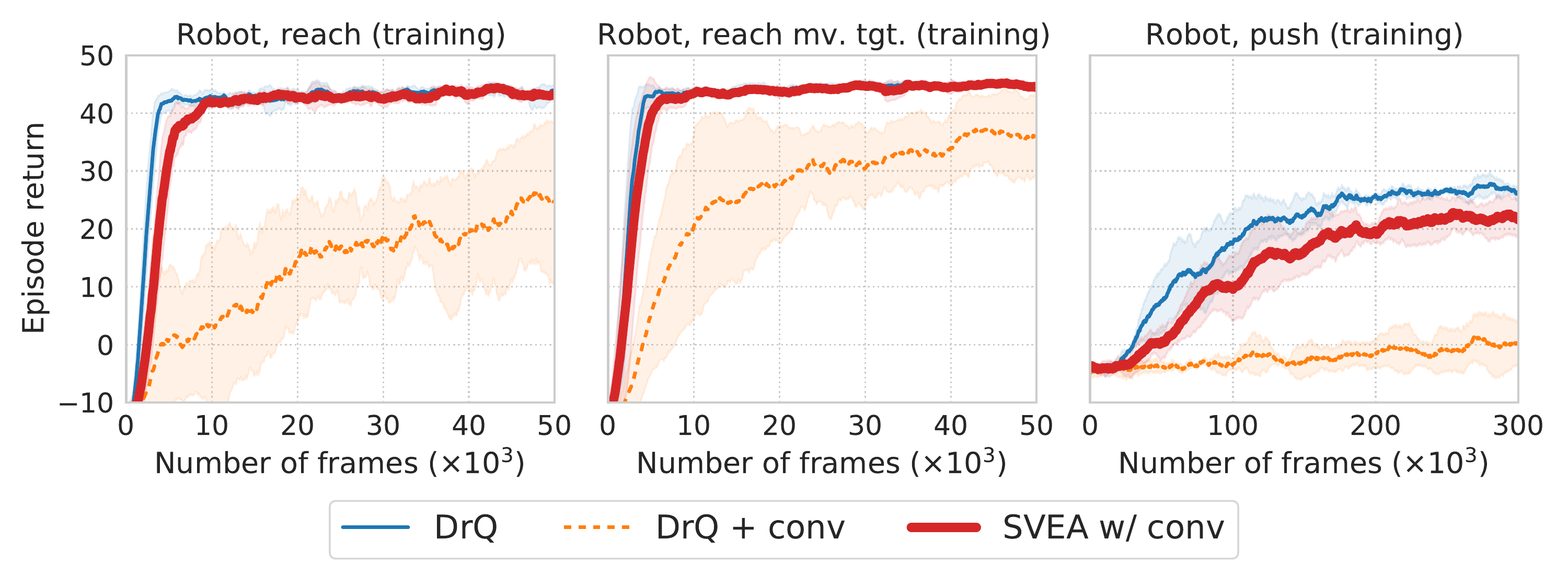}
    \vspace{-0.225in}
    \captionof{figure}{\textbf{Stability with a CNN encoder.} Training performance (episode return) of SVEA and DrQ in 3 robotic manipulation tasks. Mean and std. deviation of 5 seeds. Success rates are shown in Table \ref{tab:robot}.}
    \label{fig:robot-conv}
    \vspace{0.05in}
    \includegraphics[width=0.6\textwidth]{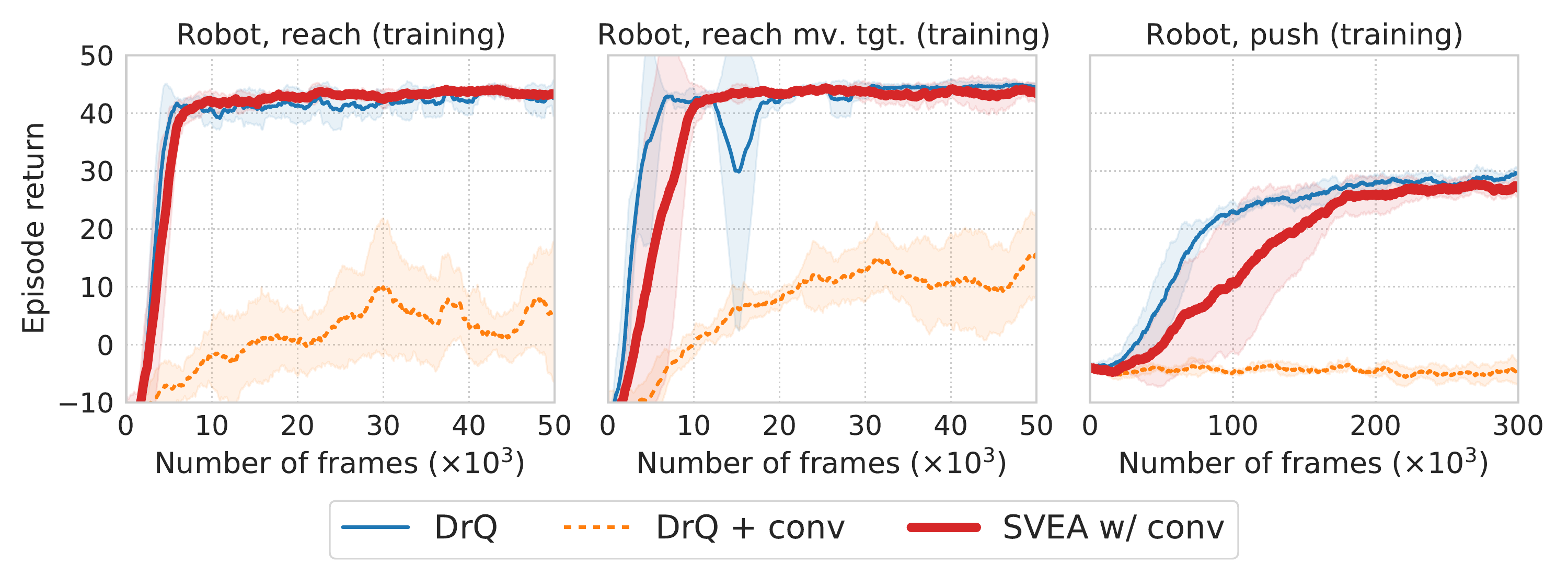}
    \vspace{-0.225in}
    \captionof{figure}{\textbf{Stability with a ViT encoder.} Training performance (episode return) of SVEA and DrQ in 3 robotic manipulation tasks. Mean and std. deviation of 5 seeds. Success rates are shown in Table \ref{tab:robot}. DrQ is especially unstable under augmentation when using a ViT encoder.}
    \label{fig:robot-vit-conv}
\end{wrapfigure}

\subsection{Robotic Manipulation}
\label{sec:experiments-robot}
\vspace{-0.075in}
We additionally consider a set of goal-conditioned robotic manipulation tasks using a simulated Kinova Gen3 arm: (i) \textit{reach}, a task in which the robot needs to position its gripper above a goal indicated by a red mark; (ii) \textit{reach moving target}, a task similar to (i) but where the robot needs to follow a red mark moving continuously in a zig-zag pattern at a random velocity; and (iii) \textit{push}, a task in which the robot needs to push a cube to a red mark. The initial configuration of gripper, object, and goal is randomized, the agent uses 2D positional control, and policies are trained using dense rewards. Observations are stacks of RGB frames with no access to state information. Training and test environments are shown in Figure \ref{fig:robot-experimental-setup}. See Appendix \ref{sec:robotic-manipulation} for further details and environment samples.

Results are shown in Figure \ref{fig:robot-conv} and Figure \ref{fig:robot-vit-conv}. For both CNN and ViT encoders, SVEA trained with \textit{conv} augmentation has similar sample efficiency and training performance as DrQ trained \textit{without} augmentation, while DrQ + conv exhibits poor sample efficiency and fails to solve the \textit{push} task. Generalization results are shown in Table \ref{tab:robot}. We find that naïve application of data augmentation has a higher success rate in test environments than DrQ, despite being less successful in the training environment, which we conjecture is because it is optimized only from augmented data. Conversely, SVEA achieves high success rates during both training and testing.

\textbf{Conclusion.} SVEA is found to greatly improve both stability and sample efficiency under augmentation, while achieving competitive generalization results. Our experiments indicate that our method scales to ViT-based architectures, and it may therefore be a promising technique for large-scale RL experiments where data augmentation is expected to play an increasingly important role.

\textbf{Broader Impact.} While our contribution aims to reduce computational cost of image-based RL, we remain concerned about the growing ecological and economical footprint of deep learning – and RL in particular – with increasingly large models such as ViT; see Appendix \ref{sec:broader-impact} for further discussion.

\textbf{Acknowledgments and Funding Transparency.} This work was supported by grants from DARPA LwLL, NSF CCF-2112665 (TILOS), NSF 1730158 CI-New: Cognitive Hardware and Software Ecosystem Community Infrastructure (CHASE-CI), NSF ACI-1541349 CC*DNI Pacific Research Platform, NSF grant IIS-1763278, NSF CCF-2112665 (TILOS), as well as gifts from Qualcomm, TuSimple and Picsart.

\bibliographystyle{plainnat}
{\small
\bibliography{main}
}

\newpage
\appendix

\section{Ablations}
\label{sec:ablations}
\vspace{-0.05in}
We ablate the design choices of SVEA and compare both training and test performance to DrQ and RAD. Results are shown in Table \ref{tab:dmc-ablations}. We find that our proposed formulation of SVEA outperforms the test performance of all other variants, and by a large margin (method \textcolor{citecolor}{2}). Using a ViT encoder (method \textcolor{citecolor}{1}) instead of a CNN further improves both the training and test performance of SVEA, whereas the test performance of DrQ \textit{decreases} by a factor of 5 when using a ViT encoder (method \textcolor{BrickRed}{7}). This indicates that ViT encoders overfit heavily to the training environment without the strong augmentation of SVEA. We observe that both DrQ and RAD are unstable under strong augmentation (method \textcolor{BrickRed}{10} and \textcolor{OliveGreen}{12}). While the test performance of DrQ does \textit{not} benefit from using a ViT encoder, we observe a slight improvement in training performance (method \textcolor{BrickRed}{7}), similar to that of SVEA.

\input{tables/dmc-ablations}

\section{Stability under Data Augmentation}
\label{sec:appendix-stability-augs}
\begin{figure}
    \centering
    \includegraphics[width=\textwidth]{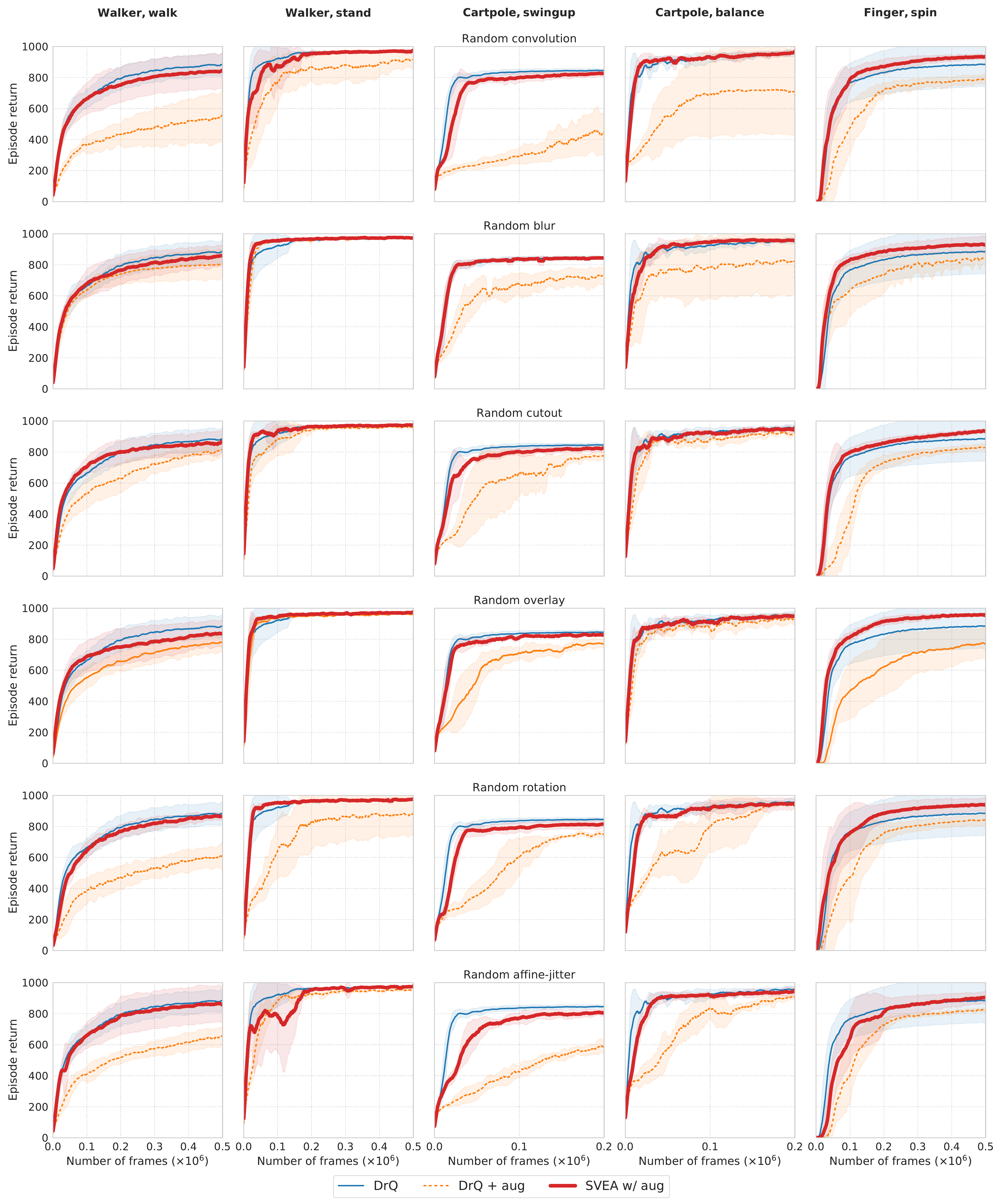}
    \caption{\textbf{Stability under data augmentation.} Training performance measured by episode return of SVEA and DrQ under 6 common data augmentations (using ConvNets). We additionally provide reference curves for DrQ without additional augmentation. Mean of 5 seeds, shaded area is $\pm1$ std. deviation. SVEA obtains similar sample efficiency to DrQ without augmentation, while the sample efficiency of \textit{DrQ + aug} is highly dependent on the task and choice of augmentation.}
    \label{fig:drq-augs-suppl}
    \vspace{-0.1in}
\end{figure}

\vspace{-0.05in}
Figure \ref{fig:drq-augs-suppl} compares the sample efficiency and stability of SVEA and DrQ under each of the 6 considered data augmentations for 5 tasks from DMControl. We observe that SVEA improves stability in all 27 instances where DrQ is impaired by data augmentation. Stability of DrQ under data augmentation is found to be highly sensitive to both the choice of augmentation and the particular task. For example, the \textit{DrQ + aug} baseline is relatively unaffected by a majority of data augmentations in the \textit{Walker, stand} task, while we observe significant instability across all data augmentations in the \textit{Cartpole, swingup} task. Our results therefore indicate that SVEA can be a highly effective method for eliminating the need for costly trial-and-error associated with application of data augmentation.

\section{Data Augmentation in RL}
\label{sec:data-augmentation}
\vspace{-0.05in}
Application of data augmentation in image-based RL has proven highly successful \citep{cobbe2018quantifying, Lee2019ASR, laskin2020reinforcement, kostrikov2020image, stooke2020atc, hansen2021softda, Raileanu2020AutomaticDA} in improving generalization by regularizing the network parameterizing the $Q$-function and policy $\pi$. However, not all augmentations are equally effective. Laskin et al. \citep{laskin2020reinforcement} and Kostrikov et al. \citep{kostrikov2020image} find small random crops and random shifts (image translations) to greatly improve sample efficiency of image-based RL, but they empirically offer no significant improvement in generalization to other environments \citep{hansen2021softda, Stone2021TheDC}. On the other hand, augmentations such as random convolution \citep{Lee2019ASR} have shown great potential in improving generalization, but is simultaneously found to cause instability and poor sample efficiency \citep{laskin2020reinforcement, hansen2021softda}. In this context, it is useful to distinguish between \textit{weak} augmentations such as small random translations that improve \textit{sample efficiency} due to their regularization, and \textit{strong} augmentations such as random convolution that improve \textit{generalization} at the expense of sample efficiency. In this work, we focus on stabilizing deep $Q$-learning under strong data augmentation with the goal of improving generalization.

\begin{figure}
    \centering
    \includegraphics[width=0.64\textwidth]{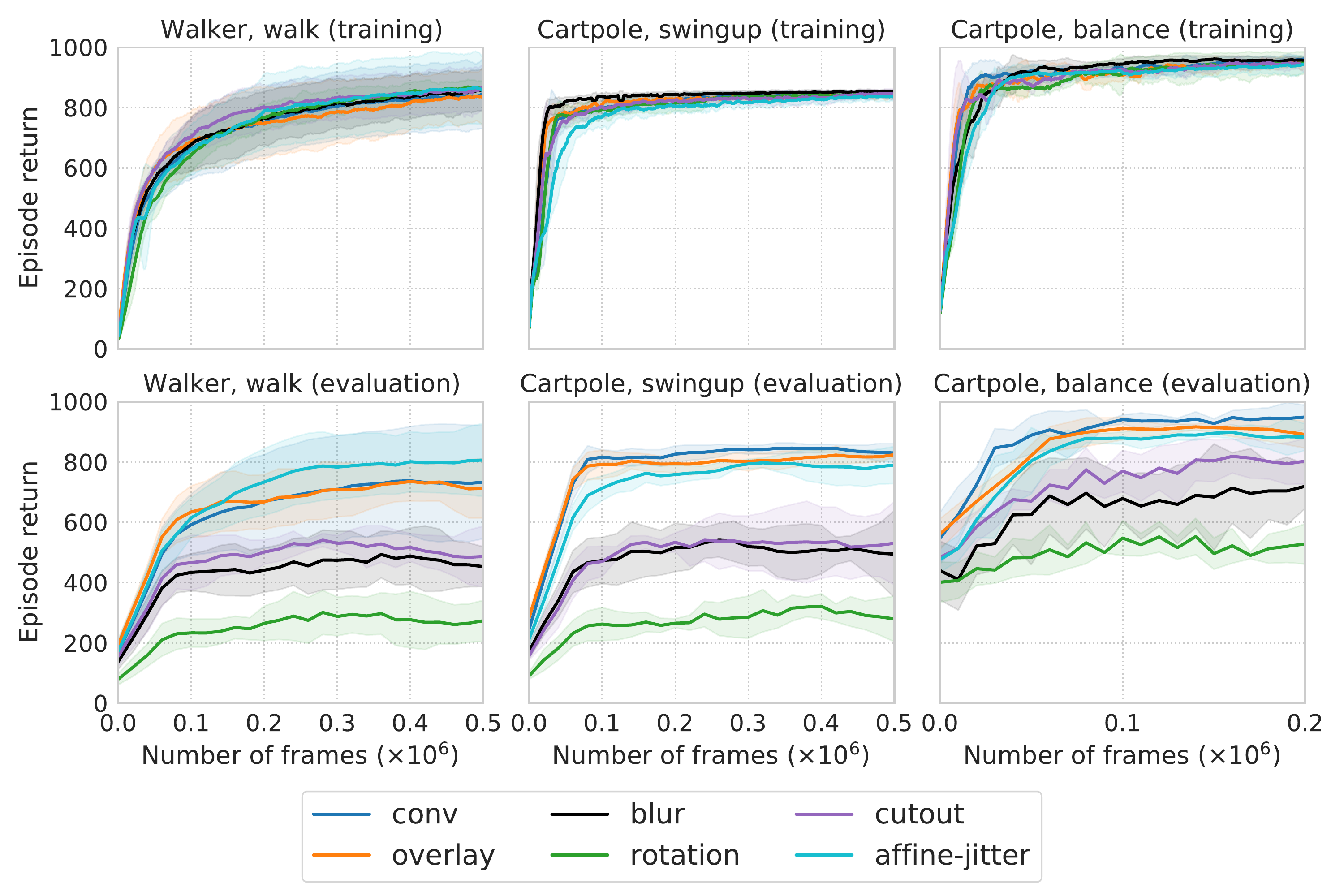}
    \vspace{-0.05in}
    \caption{\textbf{Generalization depends on the choice of data augmentation.} A comparison of SVEA implemented using each of the 6 data augmentations considered in this work (using ConvNets). SVEA exhibits comparable stability and sample efficiency for all augmentations, but generalization ability is highly dependent on the choice of augmentation. \textit{Top:} episode return on the training environment during training. \textit{Bottom:} generalization measured by episode return on the \texttt{color\_hard} benchmark of DMControl-GB. Mean of 5 seeds, shaded area is $\pm1$ std. deviation.}
    \label{fig:dmc-svea-augs-generalization}
    \vspace{-0.1in}
\end{figure}

Figure \ref{fig:dmc-svea-augs-generalization} shows training and test performance of SVEA implemented using each of the 6 data augmentations considered in this work. SVEA exhibits comparable stability and sample efficiency for all augmentations, but we find that generalization ability on the \texttt{color\_hard} benchmark of DMControl-GB is highly dependent on the choice of augmentation. Generally, we observe that augmentations such as \textit{conv}, \textit{overlay}, and \textit{affine-jitter} achieve the best generalization, but they empirically also cause the most instability in our \textit{DrQ + aug} baseline as shown in Figure \ref{fig:drq-augs-suppl}.

\begin{figure}
    \centering
    \begin{subfigure}[b]{0.48\textwidth}
        \centering
        \includegraphics[width=\textwidth]{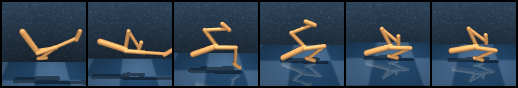}
        \caption{No augmentation (walker).}
        \vspace{0.1in}
    \end{subfigure}
    \begin{subfigure}[b]{0.48\textwidth}
        \centering
        \includegraphics[width=\textwidth]{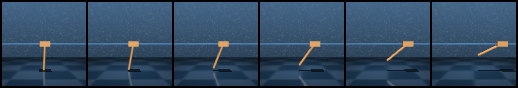}
        \caption{No augmentation (cartpole).}
        \vspace{0.1in}
    \end{subfigure}
    \begin{subfigure}[b]{0.48\textwidth}
        \centering
        \includegraphics[width=\textwidth]{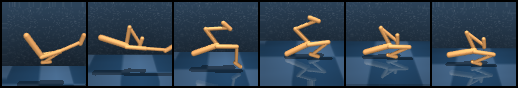}
        \caption{Random shift (walker).}
        \vspace{0.1in}
    \end{subfigure}
    \begin{subfigure}[b]{0.48\textwidth}
        \centering
        \includegraphics[width=\textwidth]{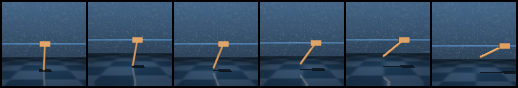}
        \caption{Random shift (cartpole).}
        \vspace{0.1in}
    \end{subfigure}
    \begin{subfigure}[b]{0.48\textwidth}
        \centering
        \includegraphics[width=\textwidth]{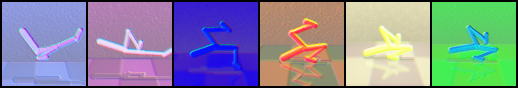}
        \caption{Random convolution (walker).}
        \vspace{0.1in}
    \end{subfigure}
    \begin{subfigure}[b]{0.48\textwidth}
        \centering
        \includegraphics[width=\textwidth]{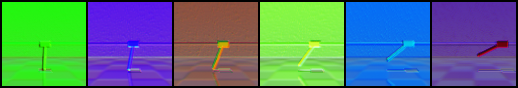}
        \caption{Random convolution (cartpole).}
        \vspace{0.1in}
    \end{subfigure}
    \begin{subfigure}[b]{0.48\textwidth}
        \centering
        \includegraphics[width=\textwidth]{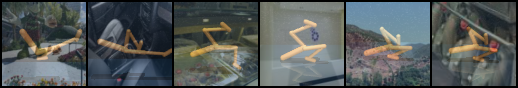}
        \caption{Random overlay (walker).}
        \vspace{0.1in}
    \end{subfigure}
    \begin{subfigure}[b]{0.48\textwidth}
        \centering
        \includegraphics[width=\textwidth]{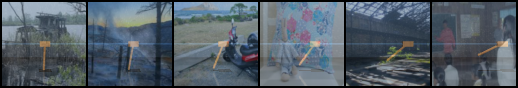}
        \caption{Random overlay (cartpole).}
        \vspace{0.1in}
    \end{subfigure}
    \begin{subfigure}[b]{0.48\textwidth}
        \centering
        \includegraphics[width=\textwidth]{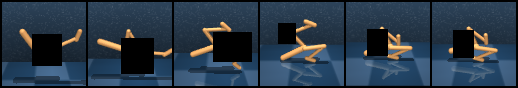}
        \caption{Random cutout (walker).}
        \vspace{0.1in}
    \end{subfigure}
    \begin{subfigure}[b]{0.48\textwidth}
        \centering
        \includegraphics[width=\textwidth]{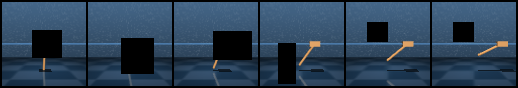}
        \caption{Random cutout (cartpole).}
        \vspace{0.1in}
    \end{subfigure}
    \begin{subfigure}[b]{0.48\textwidth}
        \centering
        \includegraphics[width=\textwidth]{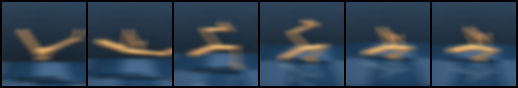}
        \caption{Random blur (walker).}
        \vspace{0.1in}
    \end{subfigure}
    \begin{subfigure}[b]{0.48\textwidth}
        \centering
        \includegraphics[width=\textwidth]{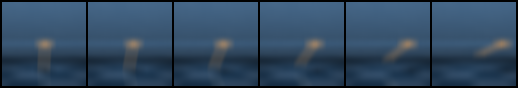}
        \caption{Random blur (cartpole).}
        \vspace{0.1in}
    \end{subfigure}
    \begin{subfigure}[b]{0.48\textwidth}
        \centering
        \includegraphics[width=\textwidth]{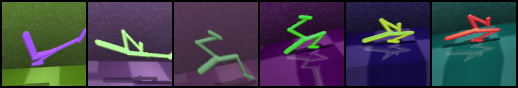}
        \caption{Random affine-jitter (walker).}
        \vspace{0.1in}
    \end{subfigure}
    \begin{subfigure}[b]{0.48\textwidth}
        \centering
        \includegraphics[width=\textwidth]{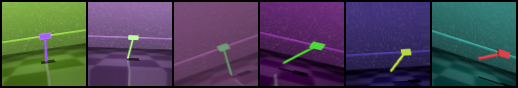}
        \caption{Random affine-jitter (cartpole).}
        \vspace{0.1in}
    \end{subfigure}
    \begin{subfigure}[b]{0.48\textwidth}
        \centering
        \includegraphics[width=\textwidth]{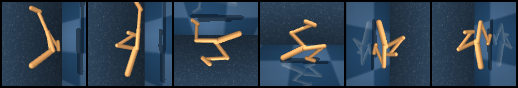}
        \caption{Random rotation (walker).}
        \vspace{0.1in}
    \end{subfigure}
    \begin{subfigure}[b]{0.48\textwidth}
        \centering
        \includegraphics[width=\textwidth]{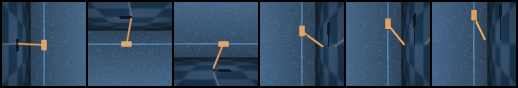}
        \caption{Random rotation (cartpole).}
        \vspace{0.1in}
    \end{subfigure}
    \caption{\textbf{Data augmentation}. Visualizations of all data augmentations considered in this study. Left column contains samples from the \textit{Walker, walk} and \textit{Walker, stand} tasks, and right column contains samples from the \textit{Cartpole, swingup} and \textit{Cartpole, balance} tasks.}
    \label{fig:data-aug-visualization}
\end{figure}

Figure \ref{fig:data-aug-visualization} provides a comprehensive set of samples for each of the data augmentations considered in this study: random \textit{shift} \citep{kostrikov2020image}, random convolution (denoted \textit{conv}) \citep{Lee2019ASR}, random \textit{overlay} \citep{hansen2021softda}, random \textit{cutout} \citep{cobbe2018quantifying}, Gaussian \textit{blur}, random \textit{affine-jitter}, and random \textit{rotation} \citep{laskin2020reinforcement}. We emphasize that the random convolution augmentation is not a convolution operation, but rather application of a randomly initialized convolutional layer as in the original proposal \citep{Lee2019ASR}. As in previous work \citep{laskin2020reinforcement, kostrikov2020image, hansen2021softda} that applies data augmentation to image-based RL, we either clip values or apply a logistic function, whichever is more appropriate, to ensure that output values remain within the $[0, 1)$ interval that unaugmented observations are normalized to. Each of the considered data augmentations are applied to the \textit{walker} and \textit{cartpole} environments and are representative of the \textit{Walker, walk}, \textit{Walker, stand}, \textit{Cartpole, swingup}, and \textit{Cartpole, balance} tasks. To illustrate the diversity of augmentation parameters associated with a given transformation, we provide a total of 6 samples for each data augmentation in each of the two environments. Note that, while random shift has been shown to improve sample efficiency in previous work, it provides very subtle randomization. Stronger and more varied augmentations such as random convolution, random overlay, and affine-jitter can be expected to improve generalization to a larger set of MDPs, but naïve application of these data augmentations empirically results in optimization difficulties and poor sample efficiency.

\begin{figure}
    \centering
    \includegraphics[width=\textwidth]{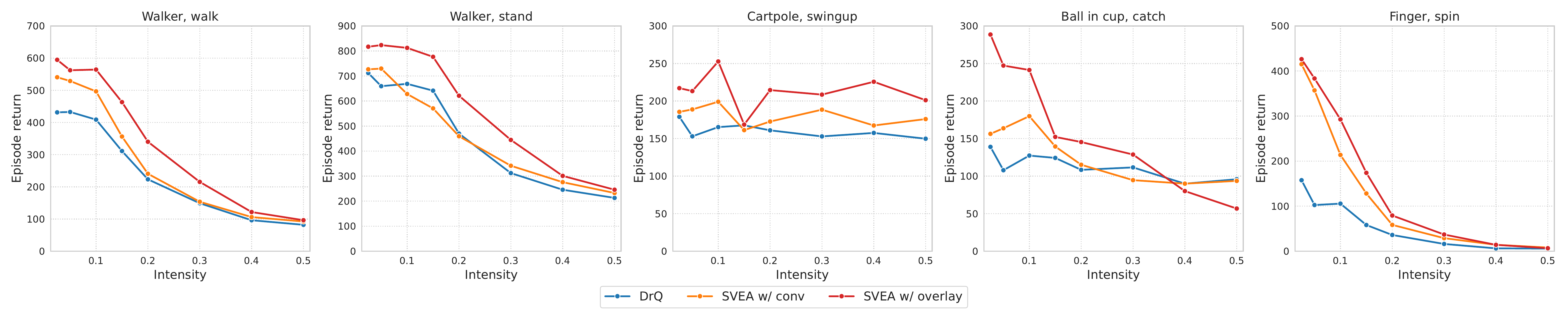}
    \vspace{-0.2in}
    \caption{\textbf{DistractingCS.} Episode return as a function of randomization intensity, for each of the 5 tasks from DMControl-GB (using ConvNets). Mean of 5 seeds. We find that the difficulty of DistractingCS varies greatly between tasks, but SVEA consistently outperforms DrQ in terms of generalization across all intensities and tasks, except for \textit{Ball in cup, catch} at the highest intensity.}
    \label{fig:dmc-dcs-individual}
    \vspace{-0.05in}
\end{figure}

\section{Choice of Base Algorithm}
\label{sec:choice-of-base-algorithm}
\begin{figure}
    \centering
    \includegraphics[width=\textwidth]{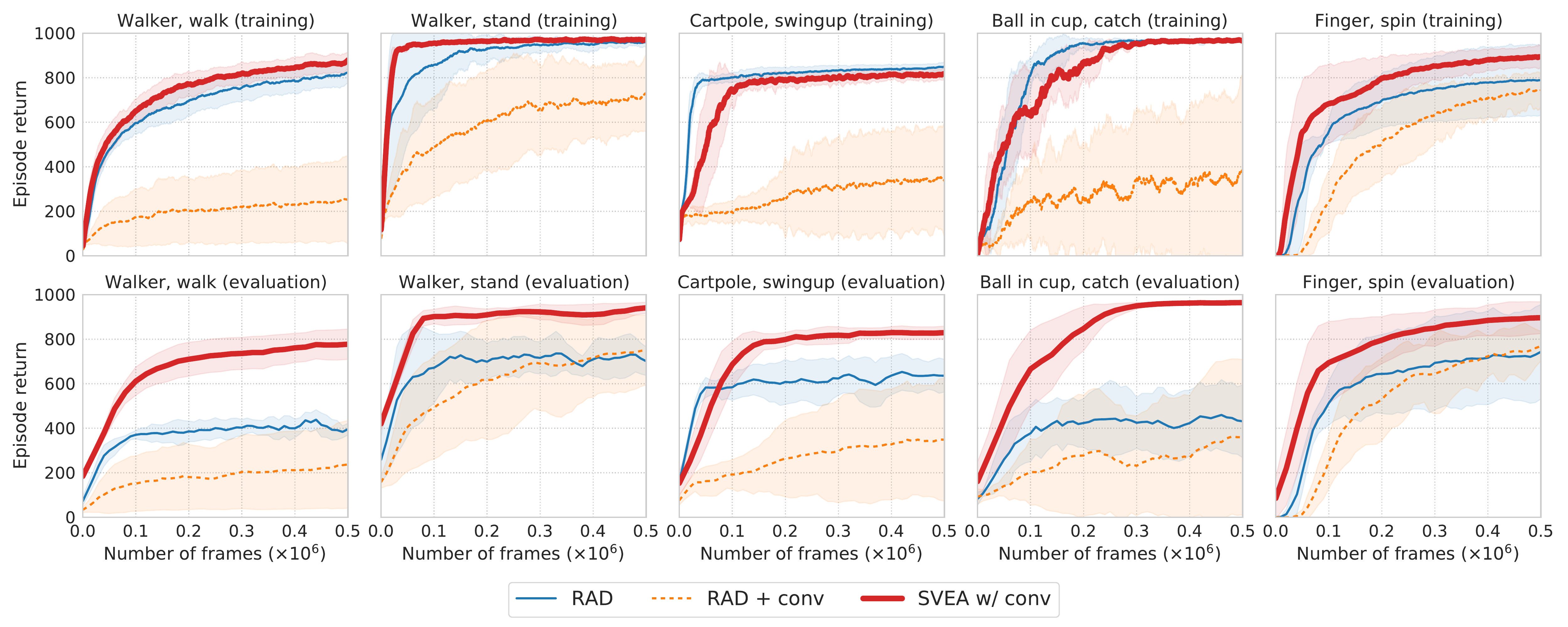}
    \vspace{-0.225in}
    \caption{\textbf{Choice of base algorithm.} We compare SVEA implemented with RAD as base algorithm to instances of RAD with and without random convolution augmentation (using ConvNets). \textit{Top:} episode return on the training environment during training. \textit{Bottom:} generalization measured by episode return on the \texttt{color\_hard} benchmark of DMControl-GB. Mean of 5 seeds, shaded area is $\pm1$ std. deviation. SVEA improves generalization in all instances.}
    \label{fig:dmc-rad-conv}
    \vspace{-0.15in}
\end{figure}

\vspace{-0.05in}
In the majority of our results, we implement SVEA using SAC as base algorithm, implemented with a random shift augmentation as proposed by DrQ \citep{kostrikov2020image}. We now consider an additional set of experiments where we instead implement SVEA using RAD \citep{laskin2020reinforcement} as base algorithm, which proposes to add a random cropping to SAC (in place of random shift). Training and test performances on \texttt{color\_hard} are shown in Figure \ref{fig:dmc-rad-conv}. We find SVEA to provide similar performance and benefits in terms of sample efficiency and generalization as when using DrQ as base algorithm. RAD likewise has similar performance to DrQ without use of strong augmentation, however, we observe that RAD generally is more unstable than DrQ when additionally using \textit{conv} augmentation, and the relative improvement of SVEA is therefore comparably larger in our RAD experiments.

\section{Test Environments}
\label{sec:test-envs}
\begin{figure}
    \centering
    \begin{subfigure}[b]{0.48\textwidth}
        \centering
        \includegraphics[width=\textwidth]{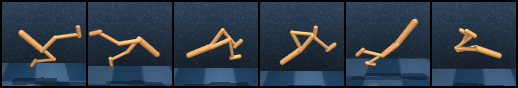}
        \caption{Training environment (walker).}
        \vspace{0.1in}
    \end{subfigure}
    \begin{subfigure}[b]{0.48\textwidth}
        \centering
        \includegraphics[width=\textwidth]{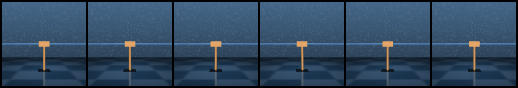}
        \caption{Training environment (cartpole).}
        \vspace{0.1in}
    \end{subfigure}
    \begin{subfigure}[b]{0.48\textwidth}
        \centering
        \includegraphics[width=\textwidth]{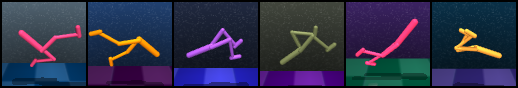}
        \caption{\texttt{color\_hard} environment (walker).}
        \vspace{0.1in}
    \end{subfigure}
    \begin{subfigure}[b]{0.48\textwidth}
        \centering
        \includegraphics[width=\textwidth]{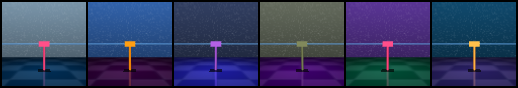}
        \caption{\texttt{color\_hard} environment (cartpole).}
        \vspace{0.1in}
    \end{subfigure}
    \begin{subfigure}[b]{0.48\textwidth}
        \centering
        \includegraphics[width=\textwidth]{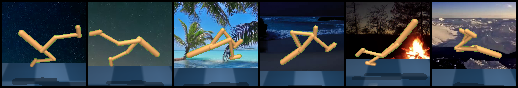}
        \caption{\texttt{video\_easy} environment (walker).}
        \vspace{0.1in}
    \end{subfigure}
    \begin{subfigure}[b]{0.48\textwidth}
        \centering
        \includegraphics[width=\textwidth]{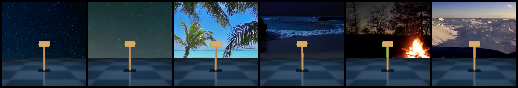}
        \caption{\texttt{video\_easy} environment (cartpole).}
        \vspace{0.1in}
    \end{subfigure}
    \begin{subfigure}[b]{0.48\textwidth}
        \centering
        \includegraphics[width=\textwidth]{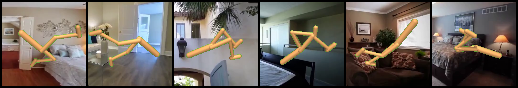}
        \caption{\texttt{video\_hard} environment (walker).}
        \vspace{0.1in}
    \end{subfigure}
    \begin{subfigure}[b]{0.48\textwidth}
        \centering
        \includegraphics[width=\textwidth]{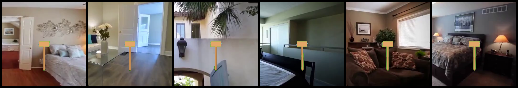}
        \caption{\texttt{video\_hard} environment (cartpole).}
        \vspace{0.1in}
    \end{subfigure}
    \begin{subfigure}[b]{0.48\textwidth}
        \centering
        \includegraphics[width=\textwidth]{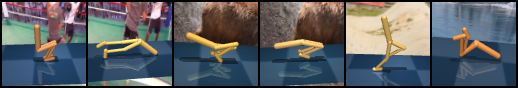}
        \caption{DistractingCS for intensity $0.1$ (walker).}
        \vspace{0.1in}
    \end{subfigure}
    \begin{subfigure}[b]{0.48\textwidth}
        \centering
        \includegraphics[width=\textwidth]{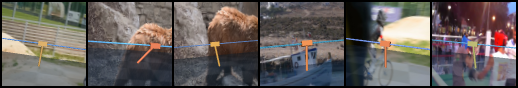}
        \caption{DistractingCS for intensity $0.1$ (cartpole).}
        \vspace{0.1in}
    \end{subfigure}
    \begin{subfigure}[b]{0.48\textwidth}
        \centering
        \includegraphics[width=\textwidth]{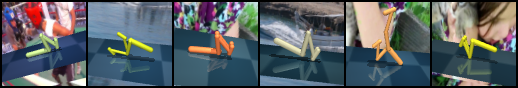}
        \caption{DistractingCS for intensity $0.2$ (walker).}
        \vspace{0.1in}
    \end{subfigure}
    \begin{subfigure}[b]{0.48\textwidth}
        \centering
        \includegraphics[width=\textwidth]{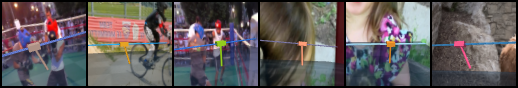}
        \caption{DistractingCS for intensity $0.2$ (cartpole).}
        \vspace{0.1in}
    \end{subfigure}
    \begin{subfigure}[b]{0.48\textwidth}
        \centering
        \includegraphics[width=\textwidth]{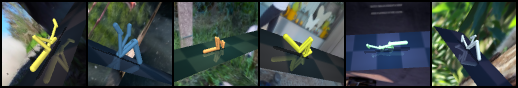}
        \caption{DistractingCS for intensity $0.5$ (walker).}
        \vspace{0.1in}
    \end{subfigure}
    \begin{subfigure}[b]{0.48\textwidth}
        \centering
        \includegraphics[width=\textwidth]{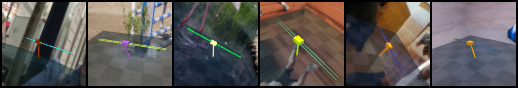}
        \caption{DistractingCS for intensity $0.5$ (cartpole).}
        \vspace{0.1in}
    \end{subfigure}
    \vspace{-0.1in}
    \caption{\textbf{Test environments}. Samples from each of the two generalization benchmarks, DMControl Generalization Benchmark \protect{\citep{hansen2021softda}} and Distracting Control Suite \protect{\citep{Stone2021TheDC}}, considered in this study. In our experiments, agents are trained in a fixed training environment with no visual variation, and are expected to generalize to novel environments of varying difficulty and factors of variation.}
    \label{fig:test-env-visualization}
    \vspace{-0.15in}
\end{figure}

\vspace{-0.05in}
Figure \ref{fig:test-env-visualization} provides visualizations for each of the two generalization benchmarks, DMControl Generalization Benchmark \citep{hansen2021softda} and Distracting Control Suite \citep{Stone2021TheDC}, used in our experiments. Agents are trained in a fixed training environment with no visual variation, and are expected to generalize to novel environments of varying difficulty and factors of variation. The \texttt{color\_hard}, \texttt{video\_easy}, and \texttt{video\_hard} benchmarks are from DMControl Generalization Benchmark, and we further provide samples from the Distracting Control Suite (DistractingCS) benchmark for intensities $I=\{0.1, 0.2, 0.5\}$. While methods are evaluated on a larger set of intensities, we here provide samples deemed representative of the intensity scale. We note that the DistractingCS benchmark has been modified to account for action repeat (frame-skip). Dynamically changing the environment at each simulation step makes the benchmark disproportionally harder for tasks that use a large action repeat, e.g. \textit{Cartpole} tasks. Therefore, we choose to modify the DistractingCS benchmark and instead update the distractors every second simulation step, corresponding to the lowest action repeat used (2, in \textit{Finger, spin}). This change affects both SVEA and baselines equally. Figure \ref{fig:dmc-dcs-individual} shows generalization results on DistractingCS for each task individually. We find that the difficulty of DistractingCS varies greatly between tasks, but SVEA consistently outperforms DrQ in terms of generalization across all intensities and tasks.

\input{tables/dmc-gb-video-hard}

\section{Additional Results on DMControl-GB}
\label{sec:dmc-gb-additional-results}
\vspace{-0.05in}
Table \ref{tab:dmc-all} contains results for the \texttt{color\_hard} and \texttt{video\_easy} generalization benchmarks from DMControl-GB. We here provide additional results for the \texttt{video\_hard} benchmark, and note that we leave out the \texttt{color\_easy} benchmark because it is already considered solved by previous work \citep{hansen2021softda, Wang2021UnsupervisedVA}. Results are shown in Table \ref{tab:dmc-video-hard}. SVEA also achieves competitive performance across all 5 tasks in the \texttt{video\_hard} benchmark.

\section{Robotic Manipulation Tasks}
\label{sec:robotic-manipulation}
\vspace{-0.05in}
We conduct experiments with a set of three goal-conditioned robotic manipulation tasks: (i) \textit{reach}, a task in which the robot needs to position its gripper above a goal indicated by a red mark; (ii) \textit{reach moving target}, a task similar to (i) but where the robot needs to follow a red mark moving continuously in a zig-zag pattern at a random velocity; and (iii) \textit{push}, a task in which the robot needs to push a cube to a red mark. We implement the tasks using MuJoCo \citep{todorov2012mujoco} for simulation, and we use a simulated Kinova Gen3 robotic arm. The initial configuration of gripper, cube, and goal is randomized, the agent uses 2D positional control, and policies are trained using dense rewards. For \textit{reach} and \textit{reach moving target}, at each time step there is a reward penalty proportional to the euclidean distance between the gripper and the goal, and there is a reward bonus of $+1$ when the distance is within a fixed threshold corresponding to the radius of the red mark. We use the same reward structure for the \textit{push} task, but use the euclidean distance between the \textit{cube} and the goal.

\begin{figure}
    \centering
    \begin{subfigure}[b]{0.48\textwidth}
        \centering
        \includegraphics[width=\textwidth]{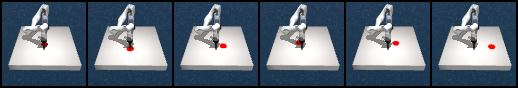}
        \caption{Training environment (reach).}
        \vspace{0.1in}
    \end{subfigure}
    \begin{subfigure}[b]{0.48\textwidth}
        \centering
        \includegraphics[width=\textwidth]{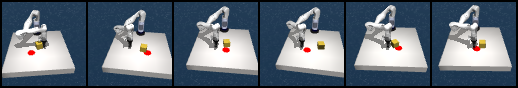}
        \caption{Training environment (push).}
        \vspace{0.1in}
    \end{subfigure}
    \begin{subfigure}[b]{0.48\textwidth}
        \centering
        \includegraphics[width=\textwidth]{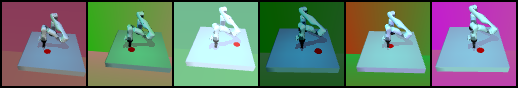}
        \caption{Test environments (reach).}
    \end{subfigure}
    \begin{subfigure}[b]{0.48\textwidth}
        \centering
        \includegraphics[width=\textwidth]{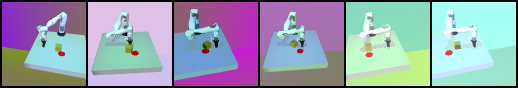}
        \caption{Test environments (push).}
    \end{subfigure}
    \caption{\textbf{Environments for robotic manipulation}. Training and test environments for our robotic manipulation experiments. Agents are trained in a fixed environment, and are expected to generalize to the unseen test environments with randomized camera pose, colors, lighting, and backgrounds.}
    \label{fig:test-env-visualization-gen3}
\end{figure}

Each episode consists of $50$ time steps, which makes $50$ an upper bound on episode return, while there is no strict lower bound. All observations are stacks of three RGB frames of size $84\times84\times3$, and the agent has no access to state information. During training, the camera position is fixed, and the camera orientation follows the gripper. During testing, the camera orientation still follows the gripper, but we additionally randomize the camera position, as well as colors, lighting, and background of the environment. Samples for the robotic manipulation tasks -- both in the training environment and the randomized test environments -- are shown in Figure \ref{fig:test-env-visualization-gen3}. We use a binary threshold for measuring task success: the episode is considered a success if the environment is in success state (i.e. either the gripper or the cube is within a fixed distance to the center of the red mark) for at least $50\%$ of all time steps in the two reaching tasks, and $25\%$ for the push task. This is to ensure that the success rate of a random policy does not get inflated by trajectories that coincidentally visit a success state for a small number of time steps, e.g. passing through the goal with the gripper.

\section{Implementation Details}
\label{sec:implementation-details}
In this section, we provide extensive implementation details for our experimental setup, including network architecture, hyperparameters, as well as design choices specific to our ViT experiments.

\textbf{Network architecture.} For experiments in DMControl \citep{deepmindcontrolsuite2018}, we adopt our network architecture from \citet{hansen2021softda}, without any changes to the architecture nor hyperparameters to ensure a fair comparison. The shared encoder $f_{\theta}$ is implemented as an 11-layer CNN encoder that takes a stack of RGB frames rendered at $84\times84\times3$ and outputs features of size $32\times 21 \times 21$, where $32$ is the number of channels and $21\times21$ are the dimensions of the spatial feature maps. All convolutional layers use $32$ filters and $3\times3$ kernels. The first convolutional layer uses a stride of 2, while the remaining convolutional layers use a stride of 1. Following previous work on image-based RL for DMControl tasks \citep{yarats2019improving, srinivas2020curl, laskin2020reinforcement, kostrikov2020image, hansen2021softda}, the shared encoder is followed by independent linear projections for the actor and critic of the Soft Actor-Critic \citep{haarnoja2018soft} base algorithm used in our experiments, and the actor and critic modules each consist of three fully connected layers with hidden dimension $1024$. Training takes approximately 24 hours on a single NVIDIA V100 GPU. For simplicity, we choose to apply the same experimental setup for robotic manipulation.

\textbf{Hyperparameters.} Whenever applicable, we adopt hyperparameters from \citet{hansen2021softda}. We detail hyperparameters relevant to our experiments in Table \ref{tab:hyperparameters-dmc}; ViT hyperparameters are discussed in the following. We use the default SVEA loss coefficients $\alpha=0.5, \beta=0.5$ in all experiments using a CNN encoder.

\begin{table}
\centering
\parbox{\textwidth}{
\caption{\textbf{Hyperparameters} used in experiments on DMControl and robotic manipulation.}
\label{tab:hyperparameters-dmc}
\vspace{0.1in}
\centering
\resizebox{0.47\textwidth}{!}{%
\begin{tabular}{@{}ll@{}}
\toprule
Hyperparameter                                                                   & Value                                                                             \\ \midrule
Frame rendering                                                                  & $84\times84\times3$                                                                     \\
Stacked frames                                                                   & 3                                                                                 \\
Random shift                                                                      & Up to $\pm4$ pixels              \\
\begin{tabular}[c]{@{}l@{}}Action repeat\\~\\~ \end{tabular}                                                                    & \begin{tabular}[c]{@{}l@{}}2 (finger)\\ 8 (cartpole)\\ 4 (otherwise)\end{tabular} \\
Discount factor $\gamma$                                                         & 0.99                                                                              \\
Episode length                                                        & 1,000                                                                                        \\
Learning algorithm                                                            & Soft Actor-Critic (SAC)                                                                    \\
Number of frames                                                            & 500,000                                                                           \\
Replay buffer size                                                               & 500,000                                                                           \\
Optimizer ($\theta$)                                                & Adam ($\beta_1=0.9, \beta_2=0.999$)                                                       \\
Optimizer ($\alpha$ of SAC)                                                      & Adam ($\beta_1=0.5, \beta_2=0.999$)                                                       \\
Learning rate ($\theta$)                                                         & 1e-3                                                      \\
Learning rate ($\alpha$ of SAC)                                                  & 1e-4                                                                              \\
Batch size                                                                    & 128 \\
SVEA coefficients                                                                    & $\alpha=0.5, \beta=0.5$ \\
$\psi$ update frequency                                                  & 2 \\
\begin{tabular}[c]{@{}l@{}}$\psi$ momentum coefficient\\~ \end{tabular}                                                                    & \begin{tabular}[c]{@{}l@{}}0.05 (encoder)\\0.01 (critic)\end{tabular}                                                                             \\ \bottomrule
\end{tabular}%
}
}
\vspace{-0.1in}
\end{table}

\textbf{RL with Vision Transformers.} We adopt a similar experimental setup for our experiments with Vision Transformers (ViT) \citep{Dosovitskiy2020AnII} as replacement for the CNN encoder used in the rest of our experiments, but with minimal changes to accommodate the new encoder. Our ViT encoder takes as input a stack of RGB frames rendered at $96\times96\times3$ (versus $84\times84\times3$ for CNN) and uses a total of $144$ non-overlapping $8\times8\times3k$ (where $k$ is the number of frames in a frame stacked observation) image patches from a spatial grid evenly placed across the image observation. All patches are projected into $128$-dimensional embeddings, and all embeddings are then forwarded as tokens for the ViT encoder. Following the original ViT implementation, we use learned positional encodings as well as a learnable \texttt{class} token. Our encoder consists of $4$ stacked Transformer \citep{Vaswani2017AttentionIA} encoders, each using Multi-Head Attention with $8$ heads. We use the ViT encoder as a drop-in replacement to CNN and optimize it jointly together with the $Q$-function using the Adam \citep{Kingma2015AdamAM} optimizer and with no changes to the learning rate. We do not pre-train the parameters of the ViT encoder, and we do not use weight decay. Similar to the CNN encoder, we find ViT to benefit from random shift augmentation and we therefore apply it by default in all methods. Training takes approximately 6 days on a single NVIDIA V100 GPU. See Table \ref{tab:hyperparameters-vit} for an overview of hyperparameters specific to the ViT encoder experiments.

\begin{table}
\centering
\parbox{0.8\textwidth}{
\caption{\textbf{Hyperparameters} used in our ViT experiments.}
\label{tab:hyperparameters-vit}
\vspace{0.1in}
\centering
\resizebox{0.475\textwidth}{!}{%
\begin{tabular}{@{}ll@{}}
\toprule
Hyperparameter                                                                    & Value                                                                             \\ \midrule
Frame rendering                                                                   & $96\times96\times3$     \\
Random shift                                                                      & Up to $\pm6$ pixels              \\
Patch size                                                                        & $8\times8\times3k$     \\
Number of patches                                                                 & $144$     \\
Embedding dimensionality                                                               & $128$                 \\
Number of layers                                                                  & $4$                 \\
Number of attention heads                                                         & $8$                 \\
Score function                                                                    & Scaled dot-product        \\
Number of frames                                                            & 300,000                                                                           \\
Replay buffer size                                                               & 300,000                                                                           \\
Batch size                                                                        & $512$                 \\
\begin{tabular}[c]{@{}l@{}}SVEA coefficients\\~\\~ \end{tabular}                                                                    & \begin{tabular}[c]{@{}l@{}}$\alpha=1, \beta=1$ (walker) \\$\alpha=1, \beta=0.5$ (push)\\$\alpha=0.5, \beta=0.5$ (otherwise)\end{tabular}\\
Number of ViT parameters                                                              & $489,600$                 \\ \bottomrule
\end{tabular}%
}
}
\end{table}

\textbf{Implementation of data augmentation in SVEA vs. previous work.} Previous work \citep{laskin2020reinforcement, kostrikov2020image, stooke2020atc, schwarzer2020data} applies augmentation to both state $\mathbf{s}^{\textnormal{aug}}_{t} = \tau(\mathbf{s}_{t}, \nu)$ and successor state $\mathbf{s}^{\textnormal{aug}}_{t+1} = \tau(\mathbf{s}_{t+1}, \nu')$ where $\nu,\nu' \sim \mathcal{V}$. As discussed in Section \ref{sec:data-augmentation}, this is empirically not found to be an issue in application of weak augmentation such as random shift \citep{kostrikov2020image}. However, previous work finds that strong augmentation, e.g. random convolution \citep{Lee2019ASR, laskin2020reinforcement}, can cause instability and poor sample efficiency. We apply random shifts in SVEA and all baselines by default, and aim to stabilize learning under strong augmentation. As such, we generically refer to observations both with and without the random shift operation as \textit{unaugmented}, and instead refer to observations as \textit{augmented} after application of one of the 6 augmentations considered in our study. While we provide the full SVEA algorithm in Algorithm \ref{alg:pseudo-code}, we here provide supplementary Python-like pseudo-code for the update rules of SVEA as well as generic off-policy actor-critic algorithms both with and without naïve application of data augmentation.

\textbf{Generic off-policy actor-critic algorithm.}
We here provide a reference implementation for a generic base algorithm from which we will implement our proposed SVEA framework for data augmentation.
\begin{codesnippet}
def update(state, action, reward, next_state):
  """Generic off-policy actor-critic RL algorithm"""
  next_action = actor(next_state)
  q_target = reward + critic_target(next_state, next_action)
  q_prediction = critic(state, action)
  
  update_critic(q_prediction, q_target)
  update_actor(state)
  update_critic_target()
\end{codesnippet}

\textbf{Naïve application of data augmentation.}
A natural way to apply data augmentation in off-policy RL algorithms is to augment both $\mathbf{s}_{t}$ and $\mathbf{s}_{t+1}$. Previous work on data augmentation in off-policy RL \citep{laskin2020reinforcement, kostrikov2020image, stooke2020atc, schwarzer2020data} follow this approach. However, this is empirically found to be detrimental to sample efficiency and stability under strong data augmentation. In the following, we generically refer to application of data augmentation as an \texttt{aug} operation.
\begin{codesnippet}
def update_aug(state, action, reward, next_state):
  """Generic off-policy actor-critic RL algorithm that uses data augmentation"""
  state = aug(state)
  next_state = aug(next_state)

  next_action = actor(next_state)
  q_target = reward + critic_target(next_state, next_action)
  q_prediction = critic(state, action)
  
  update_critic(q_prediction, q_target)
  update_actor(state)
  update_critic_target()
\end{codesnippet}

\textbf{SVEA.} Our proposed method, SVEA, does \textit{not} apply strong augmentation to $\mathbf{s}_{t+1}$ nor $\mathbf{s}_{t}$ when used for policy learning. SVEA jointly optimizes the $Q$-function over two data streams with augmented and unaugmented data, respectively, which can be implemented efficiently for $\alpha=\beta$ as in Algorithm \ref{alg:pseudo-code}. The following pseudo-code assumes $\alpha=0.5, \beta=0.5$.
\begin{codesnippet}
def update_svea(state, action, reward, next_state):
  """SVEA update for a generic off-policy actor-critic RL algorithm"""
  update_actor(state)
  next_action = actor(next_state) # [B, A]
  q_target = reward + critic_target(next_state, next_action) # [B, 1]
  
  svea_state = concatenate(state, aug(state), dim=0) # [2*B, C, H, W]
  svea_action = concatenate(action, action, dim=0) # [2*B, A]
  svea_q_target = concatenate(q_target, q_target, dim=0) # [2*B, 1]
  
  svea_q_prediction = critic(svea_state, svea_action) # [2*B, 1]
  
  update_critic(svea_q_prediction, svea_q_target)
  update_critic_target()
\end{codesnippet}
where $B$ is the batch size, $C$ is the number of input channels, $H$ and $W$ are the dimensions of observations, and $A$ is the dimensionality of the action space. For clarity, we omit hyperparameters such as the discount factor $\gamma$, learning rates, and update frequencies.

\section{Task Descriptions}
\label{sec:task-descriptions}
We experiment on tasks from DMControl \citep{deepmindcontrolsuite2018} as well as a set of robotic manipulation tasks that we implement using MuJoCo \citep{todorov2012mujoco}. DMControl tasks are selected based on previous work on both sample efficiency \citep{Hafner2019LearningLD, yarats2019improving, srinivas2020curl, Hafner2020DreamTC} and generalization \citep{hansen2021softda, Stone2021TheDC, Wang2021UnsupervisedVA}, and represent a diverse and challenging skill set in the context of image-based RL. Our set of robotic manipulation tasks are designed to represent fundamental visuomotor skills that are widely applied in related work on robot learning \citep{levine2016end, pinto2017asymmetric, ebert2018visual, nair2018visual, hansen2021deployment, young2020visual}. We here provide a unified overview of the tasks considered in our study and their properties; see Section \ref{sec:robotic-manipulation} for a detailed discussion of the robotic manipulation environment. All tasks emit observations $\mathbf{o} \in \mathbb{R}^{84\times84\times3}$ that are stacked as states $\mathbf{s} \in \mathbb{R}^{84\times84\times9}$.

\begin{itemize}
    \item\textit{Walker, walk} ($\mathbf{a} \in \mathbb{R}^{6}$). A planar walker that is rewarded for walking forward at a target velocity. Dense rewards.
    \item \textit{Walker, stand} ($\mathbf{a} \in \mathbb{R}^{6}$). A planar walker that is rewarded for standing with an upright torso at a constant minimum height. Dense rewards.
    \item \textit{Cartpole, swingup} ($\mathbf{a} \in \mathbb{R}$). Swing up and balance an unactuated pole by applying forces to a cart at its base. The agent is rewarded for balancing the pole within a fixed threshold angle. Dense rewards.
    \item \textit{Cartpole, balance} ($\mathbf{a} \in \mathbb{R}$). Balance an unactuated pole by applying forces to a cart at its base. The agent is rewarded for balancing the pole within a fixed threshold angle. Dense rewards.
    \item \textit{Ball in cup, catch} ($\mathbf{a} \in \mathbb{R}^{2}$). An actuated planar receptacle is to swing and catch a ball attached by a string to its bottom. Sparse rewards.
     \item \textit{Finger, spin} ($\mathbf{a} \in \mathbb{R}^{2}$). A manipulation problem with a planar 3 DoF finger. The task is to continually spin a free body. Sparse rewards.
     \item \textit{Robot, reach} ($\mathbf{a} \in \mathbb{R}^{2}$). A manipulation problem with a simulated Kinova Gen3 robotic arm. The task is to move the gripper to a randomly initialized goal position. Dense rewards.
     \item \textit{Robot, reach moving target} ($\mathbf{a} \in \mathbb{R}^{2}$). A manipulation problem with a simulated Kinova Gen3 robotic arm. The task is to continuously track a randomly initialized goal with the gripper. The goal moves in a zig-zag pattern at a random constant speed. Dense rewards.
     \item \textit{Robot, push} ($\mathbf{a} \in \mathbb{R}^{2}$). A manipulation problem with a simulated Kinova Gen3 robotic arm. The task is to push a cube to a goal position. All positions are randomly initialized. Dense rewards.
\end{itemize}

\section{Broader Impact}
\label{sec:broader-impact}
The discipline of deep learning -- and reinforcement learning in particular -- is rapidly evolving, which can in part be attributed to better algorithms \citep{mnih2013playing, Lillicrap2016ContinuousCW, haarnoja2018soft}, neural network architectures \citep{He2016DeepRL, Vaswani2017AttentionIA, Espeholt2018IMPALASD}, and availability of data \citep{pinto2016supersizing, Kalashnikov2018QTOptSD}, but advances are also highly driven by increased computational resources and larger models such as GPT-3 \citep{Brown2020LanguageMA} in natural language processing and ViT \citep{Dosovitskiy2020AnII} in computer vision. As a result, both computational and economic requirements for training and deploying start-of-the-art models are increasing at an unprecedented rate \citep{bender2021stochastic}. While we are concerned by this trend, we remain excited about the possibility of reusing and re-purposing large learned models (in the context of RL: policies and value functions) that learn and generalize far beyond the scope of their training environment. A greater reuse of learned policies can ultimately decrease overall computational costs, since new models may need to be trained less frequently. As researchers, we are committed to pursue research that is to the benefit of society. We strive to enable reuse of RL policies through extensive use of data augmentation, and we firmly believe that our contribution is an important step towards that goal. Our method is empirically found to reduce the computational cost (in terms of both stability, sample efficiency, and total number of gradient steps) of training RL policies under augmentation, which is an encouraging step towards learning policies that generalize to unseen environments. By extension, this promotes policy reuse, and may therefore be a promising component both for reducing costs and for improving generalization of large-scale RL that the field appears to be trending towards.

\end{document}

%% file: tables/dmc-gb.tex
\begin{table}[t]
\caption{\textbf{Comparison to state-of-the-art.} Test performance (episode return) of methods trained in a single, fixed environment and evaluated on (i) randomized colors, and (ii) natural video backgrounds from DMControl-GB. Results for CURL, RAD, PAD, and SODA are obtained from \protect{\citep{hansen2021softda}} and we report mean and std. deviation over 5 seeds. DrQ corresponds to our SAC base algorithm using random shift augmentation. SVEA matches or outperforms prior methods in all tasks considered.}
\label{tab:dmc-all}
\vspace{0.075in}
\centering
\resizebox{0.8\textwidth}{!}{%
\begin{tabular}{lcccccccc}
\toprule
DMControl-GB            & CURL & RAD & DrQ & PAD & SODA & SODA & \textbf{SVEA} & \textbf{SVEA} \\
(random colors)        & & & & & (conv) & (overlay) & (conv) & (overlay) \\\midrule
\texttt{walker,}        & $445$ & $400$ & $520$ & $468$ & $697$ & $692$ & $\mathbf{760}$ & 749 \vspace{-0.75ex} \\
\texttt{walk}           &  $\scriptstyle{\pm99}$ & $\scriptstyle{\pm61}$ & $\scriptstyle{\pm91}$ & $\scriptstyle{\pm47}$ & $\scriptstyle{\pm66}$ & $\scriptstyle{\pm68}$ & $\mathbf{\scriptstyle{\pm145}}$ & $\scriptstyle{\pm61}$ \vspace{0.75ex} \\
\texttt{walker,}        & $662$ & $644$ & $770$ & $797$ & $930$ & $893$ & $\mathbf{942}$ & $933$ \vspace{-0.75ex} \\
\texttt{stand}          & $\scriptstyle{\pm54}$ & $\scriptstyle{\pm88}$ & $\scriptstyle{\pm71}$ & $\scriptstyle{\pm46}$ & $\scriptstyle{\pm12}$ & $\scriptstyle{\pm12}$ & $\mathbf{\scriptstyle{\pm26}}$ & $\scriptstyle{\pm24}$ \vspace{0.75ex} \\
\texttt{cartpole,}      & $454$ & $590$ & $586$ & $630$ & $831$ & $805$ & $\mathbf{837}$ & 832 \vspace{-0.75ex} \\
\texttt{swingup}        & $\scriptstyle{\pm110}$ & $\scriptstyle{\pm53}$ & $\scriptstyle{\pm52}$ & $\scriptstyle{\pm63}$ & $\scriptstyle{\pm21}$ & $\scriptstyle{\pm28}$ & $\mathbf{\scriptstyle{\pm23}}$ & $\scriptstyle{\pm23}$ \vspace{0.75ex} \\
\texttt{ball\_in\_cup,} & $231$ & $541$ & $365$ & $563$ & $892$ & $949$ & $\mathbf{961}$ & $959$ \vspace{-0.75ex} \\
\texttt{catch}          & $\scriptstyle{\pm92}$ & $\scriptstyle{\pm29}$ & $\scriptstyle{\pm210}$ & $\scriptstyle{\pm50}$ & $\scriptstyle{\pm37}$ & $\scriptstyle{\pm19}$ & $\mathbf{\scriptstyle{\pm7}}$ & $\scriptstyle{\pm5}$ \vspace{0.75ex} \\
\texttt{finger,}        & $691$ & $667$ & $776$ & $803$ & $901$ & $793$ & $\mathbf{977}$ & $972$ \vspace{-0.75ex} \\
\texttt{spin}           & $\scriptstyle{\pm12}$ & $\scriptstyle{\pm154}$ & $\scriptstyle{\pm134}$ & $\scriptstyle{\pm72}$ & $\scriptstyle{\pm51}$ & $\scriptstyle{\pm128}$ & $\mathbf{\scriptstyle{\pm5}}$ & $\scriptstyle{\pm6}$ \\
\midrule
\toprule
DMControl-GB            & CURL & RAD & DrQ & PAD & SODA & SODA & \textbf{SVEA} & \textbf{SVEA} \\
(natural videos)  & & & & & (conv) & (overlay) & (conv) & (overlay) \\\midrule
\texttt{walker,}        & $556$ & $606$ & $682$ & $717$ & $635$ & $768$ & $612$ & $\mathbf{819}$ \vspace{-0.75ex}\\
\texttt{walk}           & $\scriptstyle{\pm133}$ & $\scriptstyle{\pm63}$ & $\scriptstyle{\pm89}$ & $\scriptstyle{\pm79}$ & $\scriptstyle{\pm48}$   & $\scriptstyle{\pm38}$ & $\scriptstyle{\pm144}$ & $\mathbf{\scriptstyle{\pm71}}$ \vspace{0.75ex} \\
\texttt{walker,}        & $852$ & $745$ & $873$ & $935$ & $903$ & $955$ & $795$ & $\mathbf{961}$ \vspace{-0.75ex}\\
\texttt{stand}          & $\scriptstyle{\pm75}$ & $\scriptstyle{\pm146}$ & $\scriptstyle{\pm83}$ & $\scriptstyle{\pm20}$ & $\scriptstyle{\pm56}$   & $\scriptstyle{\pm13}$ & $\scriptstyle{\pm70}$ & $\mathbf{\scriptstyle{\pm8}}$ \vspace{0.75ex}\\
\texttt{cartpole,}      & $404$ & $373$ & $485$ & $521$ & $474$ & $758$ & $606$ & $\mathbf{782}$  \vspace{-0.75ex}\\
\texttt{swingup}        & $\scriptstyle{\pm67}$ & $\scriptstyle{\pm72}$ & $\scriptstyle{\pm105}$ & $\scriptstyle{\pm76}$ & $\scriptstyle{\pm143}$  & $\scriptstyle{\pm62}$ & $\scriptstyle{\pm85}$ & $\mathbf{\scriptstyle{\pm27}}$ \vspace{0.75ex}\\
\texttt{ball\_in\_cup,} & $316$ & $481$ & $318$ & $436$ & $539$ & $\mathbf{875}$ & $659$ & $871$ \vspace{-0.75ex}\\
\texttt{catch}          & $\scriptstyle{\pm119}$ & $\scriptstyle{\pm26}$ & $\scriptstyle{\pm157}$ & $\scriptstyle{\pm55}$ & $\scriptstyle{\pm111}$ & $\mathbf{\scriptstyle{\pm56}}$ & $\scriptstyle{\pm110}$ & $\scriptstyle{\pm106}$ \vspace{0.75ex}\\
\texttt{finger,}        & $502$ & $400$ & $533$ & $691$ & $363$ & $695$ & $764$ & $\mathbf{808}$ \vspace{-0.75ex}\\
\texttt{spin}           & $\scriptstyle{\pm19}$ & $\scriptstyle{\pm64}$ & $\scriptstyle{\pm119}$ & $\scriptstyle{\pm80}$ & $\scriptstyle{\pm185}$ & $\scriptstyle{\pm97}$ & $\scriptstyle{\pm86}$ & $\mathbf{\scriptstyle{\pm33}}$ \\
\bottomrule
\end{tabular}
}
\vspace{-0.1in}
\end{table}

%% file: tables/dmc-ablations.tex
\begin{table}[h]
\vspace{-0.075in}
\caption{\textbf{Ablations.} We vary the following choices: (i) architecture of the encoder; (ii) our proposed objective $\mathcal{L}^{\textbf{SVEA}}_{Q}$ as opposed to $\mathcal{L}_{Q}$ or a \textit{mix-all} objective that uses two data-streams for both $Q$-predictions, $Q$-targets, and $\pi$; (iii) using strong augmentation (\textit{conv}) in addition to the random shift augmentation used by default (abbreviated as \textit{Str. aug.}); and (iv) whether the target is augmented or not (abbreviated as \textit{Aug. tgt.}). We report mean episode return in the training and test environments (\texttt{color\_hard}) of the \textit{Walker, walk} task. Method \textcolor{citecolor}{1} and \textcolor{citecolor}{2} are the default formulations of SVEA using ViT and CNN encoders, respectively, method \textcolor{BrickRed}{7} and \textcolor{BrickRed}{8} are the default formulations of DrQ using Vit and CNN encoders, respectively, and method \textcolor{OliveGreen}{11} is the default formulation of RAD that uses a random crop augmentation and is implemented using a CNN encoder. Mean and std. deviation. of 5 seeds.}
\label{tab:dmc-ablations}
\vspace{0.1in}
\centering
\resizebox{0.75\textwidth}{!}{%
\begin{tabular}{llcccccc}
\toprule
 & Method  & Arch. & Objective & Str. aug. & Aug. tgt. & Train return & Test return \\\midrule
\color{citecolor}{1} & \textbf{SVEA} & ViT   & SVEA    & \Checkmark      &  \XSolidBrush   & $918\scriptstyle{\pm57}$    & $\mathbf{877\scriptstyle{\pm54}}$ \\
\color{citecolor}{2} & \color{Gray}{$-$} & CNN   & SVEA    & \Checkmark      &  \XSolidBrush   & $833\scriptstyle{\pm91}$     & $760\scriptstyle{\pm145}$ \\
\color{citecolor}{3} & \color{Gray}{$-$}  & \textcolor{Gray}{CNN}  & \textcolor{Gray}{SVEA}  & \color{Gray}{\Checkmark}      & \color{Gray}{\Checkmark}      & $872\scriptstyle{\pm53}$      & $605\scriptstyle{\pm148}$  \\
\color{citecolor}{4} & \color{Gray}{$-$} & \textcolor{Gray}{CNN}    & \textcolor{Gray}{mix-all}     & \color{Gray}{\Checkmark}      & \color{Gray}{\Checkmark}      & $\mathbf{927\scriptstyle{\pm24}}$      & $599\scriptstyle{\pm214}$   \\
\color{citecolor}{5} & \color{Gray}{$-$} & \textcolor{Gray}{CNN}   & \textcolor{Gray}{$Q$}    & \color{Gray}{\Checkmark}         & \color{Gray}{\XSolidBrush}      & $596\scriptstyle{\pm55}$       & $569\scriptstyle{\pm139}$  \\
\color{citecolor}{6} & \color{Gray}{$-$} & \textcolor{Gray}{CNN}   & \textcolor{Gray}{$Q$}    & \color{Gray}{\XSolidBrush}         & \color{Gray}{\XSolidBrush}      & $771\scriptstyle{\pm317}$      & $498\scriptstyle{\pm196}$      \\
\cmidrule(lr){1-8}
\color{BrickRed}{7} & \textbf{DrQ} & ViT    & $Q$      & \XSolidBrush      & \Checkmark      & $\mathbf{920\scriptstyle{\pm36}}$    & $94\scriptstyle{\pm18}$ \\
\color{BrickRed}{8} & \color{Gray}{$-$} & CNN    & $Q$      & \XSolidBrush      & \Checkmark      & $892\scriptstyle{\pm65}$    & $520\scriptstyle{\pm91}$ \\
\color{BrickRed}{9} & \color{Gray}{$-$} & \textcolor{Gray}{ViT}    & \textcolor{Gray}{$Q$}     & \color{Gray}{\Checkmark}      & \color{Gray}{\Checkmark}      & $286\scriptstyle{\pm225}$    & $470\scriptstyle{\pm67}$  \\
\color{BrickRed}{10} & \color{Gray}{$-$} & \textcolor{Gray}{CNN}  & \textcolor{Gray}{$Q$}     & \color{Gray}{\Checkmark}      & \color{Gray}{\Checkmark}      & $560\scriptstyle{\pm158}$      & $\mathbf{569\scriptstyle{\pm139}}$     \\
\cmidrule(lr){1-8}
\color{OliveGreen}{11} & \textbf{RAD} & CNN   & $Q$      & \XSolidBrush      & \Checkmark      & $\mathbf{883\scriptstyle{\pm23}}$     & $\mathbf{400\scriptstyle{\pm61}}$ \\
\color{OliveGreen}{12} & \color{Gray}{$-$} & \textcolor{Gray}{CNN} & \textcolor{Gray}{$Q$}     & \color{Gray}{\Checkmark}      & \color{Gray}{\Checkmark}      & $260\scriptstyle{\pm201}$        & $246\scriptstyle{\pm184}$      \\
\bottomrule
\end{tabular}
}
\vspace{-0.075in}
\end{table}

%% file: tables/dmc-gb-video-hard.tex
\begin{table}[h]
\caption{\textbf{Comparison to state-of-the-art.} Test performance (episode return) of methods trained in a fixed environment and evaluated on the \texttt{video\_hard} benchmark from DMControl-GB. In this setting, the entirety of the floor and background is replaced by natural videos; see Figure \ref{fig:test-env-visualization} for samples. Results for CURL, RAD, PAD, and SODA are obtained from \protect{\citet{hansen2021softda}} and we report mean and std. deviation of 5 runs. We compare SVEA and SODA using the \textit{overlay} augmentation, since this is the augmentation for which the strongest results are reported for SODA in \protect{\citet{hansen2021softda}}. SVEA achieves competitive results in all tasks considered, though there is still some room for improvement before the benchmark is saturated.}
\label{tab:dmc-video-hard}
\vspace{0.1in}
\centering
\resizebox{0.7\textwidth}{!}{%
\begin{tabular}{lcccccc}
\toprule
DMControl-GB            & CURL & RAD & DrQ & PAD & SODA & \textbf{SVEA} \\
(\texttt{video\_hard})        & & & & & (overlay) & (overlay) \\\midrule
\texttt{walker,}        & $58$ & $56$ & $104$ & $93$ & $\mathbf{381}$ & $377$ \vspace{-0.75ex} \\
\texttt{walk}           &  $\scriptstyle{\pm18}$ & $\scriptstyle{\pm9}$ & $\scriptstyle{\pm22}$ & $\scriptstyle{\pm29}$ & $\mathbf{\scriptstyle{\pm72}}$ & $\scriptstyle{\pm93}$ \vspace{0.75ex} \\
\texttt{walker,}        & $45$ & $231$ & $289$ & $278$ & $771$ & $\mathbf{834}$ \vspace{-0.75ex} \\
\texttt{stand}          &  $\scriptstyle{\pm5}$ & $\scriptstyle{\pm39}$ & $\scriptstyle{\pm49}$ & $\scriptstyle{\pm72}$ & $\scriptstyle{\pm83}$ & $\mathbf{\scriptstyle{\pm46}}$ \vspace{0.75ex} \\
\texttt{cartpole,}      & $114$ & $110$ & $138$ & $123$ & $\mathbf{429}$ & $393$ \vspace{-0.75ex} \\
\texttt{swingup}       &  $\scriptstyle{\pm15}$ & $\scriptstyle{\pm16}$ & $\scriptstyle{\pm9}$ & $\scriptstyle{\pm24}$ & $\mathbf{\scriptstyle{\pm64}}$ & $\scriptstyle{\pm45}$ \vspace{0.75ex} \\
\texttt{ball\_in\_cup,} & $115$ & $97$ & $92$ & $66$ & $327$ & $\mathbf{403}$ \vspace{-0.75ex} \\
\texttt{catch}          &  $\scriptstyle{\pm33}$ & $\scriptstyle{\pm29}$ & $\scriptstyle{\pm23}$ & $\scriptstyle{\pm61}$ & $\scriptstyle{\pm100}$ & $\mathbf{\scriptstyle{\pm174}}$ \vspace{0.75ex} \\
\texttt{finger,}        & $27$ & $34$ & $71$ & $56$ & $302$ & $\mathbf{335}$ \vspace{-0.75ex} \\
\texttt{spin}           &  $\scriptstyle{\pm21}$ & $\scriptstyle{\pm11}$ & $\scriptstyle{\pm45}$ & $\scriptstyle{\pm18}$ & $\scriptstyle{\pm41}$ & $\mathbf{\scriptstyle{\pm58}}$ \vspace{0.75ex}\\
\bottomrule
\end{tabular}
}
\end{table}